%% file: main.tex
\documentclass[runningheads]{llncs}

\usepackage{eccv}

\usepackage{eccvabbrv}

\usepackage{graphicx}
\usepackage{booktabs}

\usepackage[accsupp]{axessibility}  

\usepackage{url}

\usepackage[dvipsnames]{xcolor}

\usepackage{multirow} 
\usepackage{amsmath} 
\usepackage{bbm} 
\usepackage{siunitx} 
\usepackage{kotex} 
\usepackage{array,graphicx}
\usepackage{float}
\usepackage{mathtools}
\usepackage{bbm}
\usepackage{hhline}
\usepackage{boldline}
\usepackage{tabularx}
\usepackage{comment}
\usepackage{color}
\usepackage{amssymb}
\usepackage{booktabs}
\usepackage{multicol}
\usepackage{microtype}      
\usepackage{xcolor}         
\usepackage{algpseudocode}
\usepackage{pifont}
\usepackage{soul} 
\usepackage{subcaption}
\usepackage{wrapfig}
\usepackage{lipsum}
\usepackage{colortbl}
\usepackage{nicematrix}
\usepackage{dsfont}
\usepackage{caption}
\usepackage[accsupp]{axessibility}
\usepackage{algorithm}
\usepackage{enumitem}
\usepackage[table]{xcolor} 
\usepackage{tikz}

\definecolor{bestred}{HTML}{FFC3C3}
\definecolor{secondorange}{HTML}{FFD9B2}
\definecolor{thirdyellow}{HTML}{FFFFB2}

\newcommand{\best}[1]{\cellcolor{bestred}\textbf{#1}}
\newcommand{\second}[1]{\cellcolor{secondorange}\textbf{#1}}
\newcommand{\third}[1]{\cellcolor{thirdyellow}\textbf{#1}}

\newcommand{\hlcaption}[2]{%
    \setlength{\fboxsep}{1.5pt}
    \colorbox{#1}{#2}
}

\definecolor{secondorange}{HTML}{FFD9B2}
\newcommand{\xmark}{\ding{55}}%
\usepackage{amsfonts}

\usepackage{hyperref}

\usepackage{orcidlink}

\begin{document}

\title{Selective Synergistic Learning \\for Video Object-Centric Learning} 

\titlerunning{SSync}


\author{
WonJun Moon\inst{1}$^{\dagger}$\orcidlink{0000-0003-2805-0926}
\and
Jae-Pil Heo\inst{2}\thanks{Corresponding author}\orcidlink{0000-0001-9684-7641}
}

\authorrunning{WJ.\ Moon et al.}

\institute{
KAIST, South Korea\\
\and
Sungkyunkwan University, South Korea\\
}
\begingroup
\renewcommand\thefootnote{$\dagger$}
\footnotetext{Work done at Sungkyunkwan University.}
\endgroup


\authorrunning{WJ.~Moon et al.}

\maketitle
\input{sec/0_abstract}
\input{sec/1_introduction}
\input{sec/2_related_work}
\input{sec/3_method}

\input{sec/4_experiment}

\input{sec/5_conclusion}

\input{sec/appendix}

\clearpage
\bibliographystyle{unsrt}    
\bibliography{main}

\end{document}

%% file: sec/0_abstract.tex
\begin{abstract}
Typical video object-centric learning~(VOCL) approaches employ slot-based frameworks that rely on reconstruction-driven encoder–decoder architectures, where learning is mediated by two spatial maps: attention maps from the encoder and object maps from the decoder.
As these two distinct maps exhibit different properties, a recent dense alignment strategy attempted to reconcile this discrepancy by enforcing agreement across all spatio-temporal patches via contrastive learning. 
However, this indiscriminate alignment inadvertently propagates the inherent weaknesses of each module, such as noisy encoder predictions and blurred decoder boundaries.
Moreover, computing dense similarities across all pairs incurs a computational cost quadratic in the total number of spatio-temporal patches, severely limiting scalability.
Motivated by this, we propose Selective Synergistic Learning~(SSync).
Instead of exhaustive patch-to-patch alignment, SSync prevents error propagation by selectively distilling only the most reliable cues: leveraging the encoder strictly for boundary refinement and the decoder for interior denoising.
This is realized via a pseudo-labeling with linear complexity, eliminating the need for quadratic spatial comparisons.
Also, to prevent the reinforcement of architectural biases like slot redundancy, we introduce a transitive pseudo-label merging that consolidates overlapping slots based on spatio-temporal activation consistency.
Extensive studies demonstrate that SSync improves decomposition quality and serves as a versatile, plug-and-play module while also exhibiting exceptional robustness to slot configurations.
Code is available at \href{github.com/wjun0830/SSync}{github.com/wjun0830/SSync}.
\keywords{Video Object-Centric Learning \and Object Discovery}
\end{abstract}

%% file: sec/1_introduction.tex
\section{Introduction}
Object-centric learning is a fundamental paradigm for structured visual reasoning, enabling downstream applications such as object editing, generation, and scene understanding~\cite{slotvae, wu2023slotdiffusion, ocslotdiffusion, slotvlm, jeon2025masking, 3dgraphllm, chatscene}.
By decomposing scenes into object representations, object-centric frameworks provide an interpretable and modular representation space that facilitates reasoning beyond pixel-level perception~\cite{uni3d}. 
Among them, slot attention~\cite{slotattention} has emerged as a dominant method, grouping patch features into latent slots and reconstructing the input through a decoder that implicitly produces object maps. 
It has recently been extended to videos, leveraging temporal cues to achieve consistent object discovery and tracking across frames~\cite{videosaur, steve, savi, elsayed2022savi++, srl, slotcurri}.

Despite their empirical success, slot architectures suffer from a fundamental structural mismatch.
Learning is mediated by two spatial maps: the encoder’s attention maps and the decoder’s reconstructed object maps. 
Ideally, these maps should be logically consistent, as both branches are jointly optimized through a shared reconstruction objective. 
However, in practice, they exhibit distinct inductive biases. 
The encoder, often built upon high-capacity vision backbones~\cite{Dinov2}, produces spatially sharp yet noisy assignments, exhibiting high-frequency sensitivity. 
In contrast, the decoder, typically implemented as a lightweight MLP, imposes a low-frequency spatial prior, yielding smooth and temporally coherent but blurry object boundaries~\cite{srl}.
This structural asymmetry induces persistent misalignment between patch assignments and rendering regions, resulting in unstable slot semantics and fragmented object decomposition.

\begin{figure}[t]
\centering
\vspace{-0.25cm}
\includegraphics[width=0.76\columnwidth]{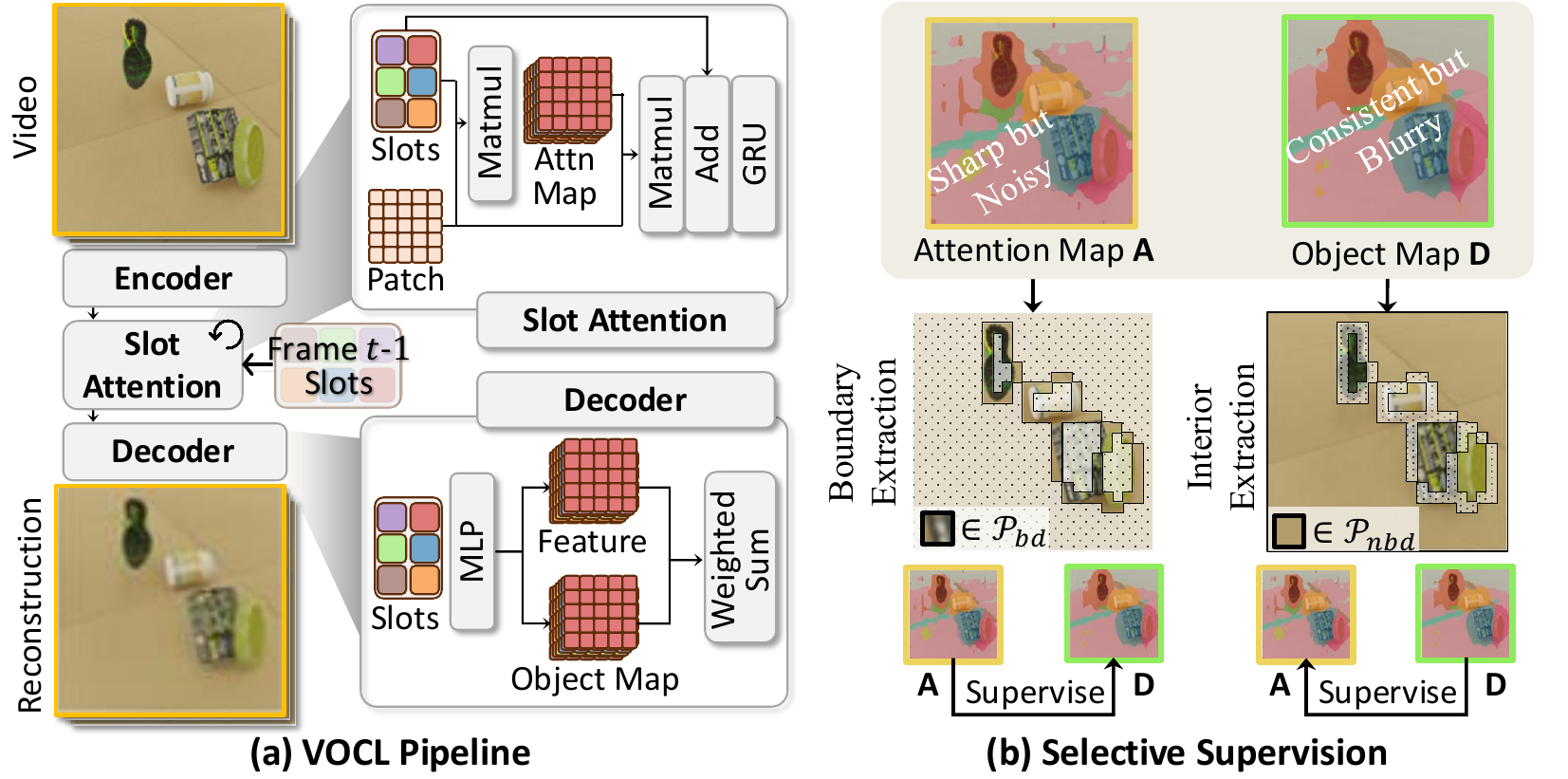} 
\vspace{-0.25cm}
\caption{
\textbf{Overall flow and motivation.}
(a) Video frames are sequentially processed, where slots are recurrently updated based on the previous frame’s slots. 
(b) Slot attention map captures sharp boundaries but contains noise, while the decoder object map offers a consistent representation but suffers from blurry edges. 
To leverage their complementary strengths, SSync reciprocally distills the sharp boundary cues from the attention map and the consistent semantics of the decoder object map.
}
\label{fig1.motiv}
\vspace{-0.45cm}
\end{figure}

A straightforward solution is to enforce consistency between these two spatial maps. 
Recently, SRL~\cite{srl} attempts to bridge this gap through dense contrastive alignment across all spatio-temporal patches. 
While conceptually appealing, we argue that dense alignment is fundamentally misaligned with the heterogeneous inductive biases of the encoder and decoder since it treats every patch as an equally reliable teacher.
Consequently, it inadvertently propagates the inherent failure modes of each branch. 
Also, computing similarities across all spatio-temporal patches incurs a quadratic memory cost, severely limiting its applicability to long sequences or high-resolution videos.

In this work, we introduce a selective alignment principle: mutual supervision should occur only where each branch provides reliable cues.
A key insight is that, once reliable regions are correctly identified, the alignment mechanism itself can be remarkably simple.
We introduce Selective Synergistic Learning~(SSync), a selective mutual distillation framework that filters supervision at the patch level instead of enforcing global agreement.
Concretely, we exploit the encoder’s boundary sensitivity to refine decoder masks at object boundaries, while leveraging the decoder’s spatial coherence to denoise encoder assignments within interior regions.
Specifically, we exploit local consistency cues to discern object boundaries within the encoder and coherent interior regions within the decoder, and formulate an efficient supervisory signal through pseudo-labeling.
By aligning only in these reliable regions, SSync mitigates error propagation inherent in dense objectives while substantially reducing memory complexity.
Our contributions are: (1) \textbf{Reliability-based selective distillation.} We identify the spatial complementarity between encoder and decoder expertise and reformulate mutual learning as a selective cross-distillation framework, (2) \textbf{Redundancy control via transitive merging.} To stabilize on-the-fly pseudo-labeling during optimization, we introduce a transitive merging strategy, and (3) \textbf{Strong performance \& practical generalization.} SSync achieves state-of-the-art results across VOCL benchmarks, improves memory efficiency relative to dense alignment, and remains robust to varying slot configurations.

%% file: sec/2_related_work.tex
\section{Related Work}

\subsection{Object-Centric Representation Learning}
Object-centric learning seeks to decompose scenes into discrete structural entities~\cite{seitzer2023bridging, liu2025metaslot, zhao2025multiscale}.
Slot Attention~\cite{slotattention} established a foundational framework by iteratively grouping spatial features into latent slots and reconstructing the image via a shared decoder. 
This encoder–decoder formulation enables unsupervised object discovery while maintaining permutation invariance over slots.

Recent works have extended slot-based learning to dynamic visual scenes~\cite{randsf}. SAVi~\cite{savi} and SAVi++~\cite{elsayed2022savi++} leveraged motion cues such as optical flow and depth to promote temporal consistency. 
STEVE~\cite{steve} incorporated transformer-based models to capture complex object interactions. 
Videosaur~\cite{videosaur} introduced objectives that encourage grouping of motion-consistent regions, and SlotContrast~\cite{slotcontrast} enhanced slot discriminability via temporal contrastive learning.
While these significantly improve object discovery in videos, they predominantly rely on a reconstruction objective, which implicitly assumes a convergence between encoder attention and decoder object maps. 
In practice, however, these two representations exhibit distinct inductive biases, leading to spatial discrepancies.

To address this discrepancy, SRL~\cite{srl} introduced a mutual refinement process via a dense alignment strategy, employing a contrastive objective across all spatio-temporal patches. 
While this reduces inconsistency, it overlooks the unique strengths and inherent limitations of each branch by uniformly enforcing agreement across all regions. This indiscriminate alignment inevitably conflates complementary strengths with architectural weaknesses, all while incurring prohibitive memory costs.
In contrast, we explicitly identify the expertise of each branch and instantiate synergy through selective cross-distillation, thereby preventing error propagation and ensuring linear scalability.

\subsection{Pseudo-Labeling}
Pseudo-labeling has been widely adopted in semi-supervised and self-supervised learning to exploit high-confidence predictions as supervisory signals~\cite{pseudolabel1, pseudolabel2, pseudolabel3, pseudolabel4, dividemix, zhang2021flexmatch}.
Self-training approaches such as Noisy Student~\cite{xie2020self} and Meta Pseudo Labels~\cite{pham2021meta} also demonstrate that iterative refinement of pseudo-labels can significantly improve representation quality. 

However, directly applying pseudo-labeling to slot learning introduces unique challenges. 
Since pseudo-labels are generated on-the-fly from the model's own predictions, they are inherently unstable; early training errors can easily propagate via pseudo-labeling, making the framework highly vulnerable to noise.
To resolve this, we introduce a transitive pseudo-label merging that analyzes spatio-temporal activation overlap to detect redundancy and consolidate slot identities globally. 
This redundancy-aware refinement stabilizes selective supervision and prevents feedback loops caused by imperfect on-the-fly pseudo-labels.


%% file: sec/3_method.tex
\section{Selective Synergistic Learning}
\label{sec:method}

\subsection{Preliminaries}
\textbf{Video Slot Learning and Spatial Maps.} We consider an input video represented as a grid of patch tokens.
Let $z_{t,p}$ denote the feature of patch $p \in \{1,\dots,P\}$ at time $t \in \{1,\dots,T\}$, where $P=H\times W$.
A slot attention encoder predicts $S$ slot prototypes. By computing the dot product between the projected patch queries $\mathbf{q}_{t,p}$~(derived from $z_{t,p}$) and the slot keys $\mathbf{k}_{t,s}$, the encoder generates an encoder attention map over slots for each patch:
\begin{equation}
\mathbf{A}_{t,p} = \mathrm{Softmax}\big(\mathbf{q}_{t,p}^\top \mathbf{k}_{t,1:S}\big),
\label{eq:attn_map}
\end{equation}
where $\mathbf{A}_{t,p,s}$ is the probability that patch $z_{t,p}$ is explained by slot $s$.
A decoder reconstructs the input and produces a decoder object map $\mathbf{D} \in \mathbb{R}^{T \times P \times S}$, which is normalized over the slot dimension such that $\sum_{s=1}^S \mathbf{D}_{t,p,s} = 1$ for all $t$ and $p$.
Both maps provide patch-to-slot assignments but are generated by different modules and thus exhibit different error characteristics.

\textbf{Bias Misalignment in Encoder-Decoder Maps.}
Ideally, the attention map $\mathbf{A}$ and the object map $\mathbf{D}$ should be consistent, since they are optimized jointly under reconstruction-driven learning.
In practice, however, $\mathbf{A}$ and $\mathbf{D}$ are persistently misaligned since they inherit distinct spatial characteristics from different modules.
The encoder typically produces spatially sharp attention maps but is vulnerable to noisy assignments, while the decoder tends to yield spatio-temporally coherent maps but with blurred object boundaries.
This heterogeneity implies that reliability is region-dependent: boundary regions benefit from the encoder's sharpness, whereas interior regions benefit from the decoder's coherence.

\textbf{Limitations of Dense Alignment.}
\begin{figure}[t]
\centering
\vspace{-0.25cm}
\includegraphics[width=0.93\columnwidth]{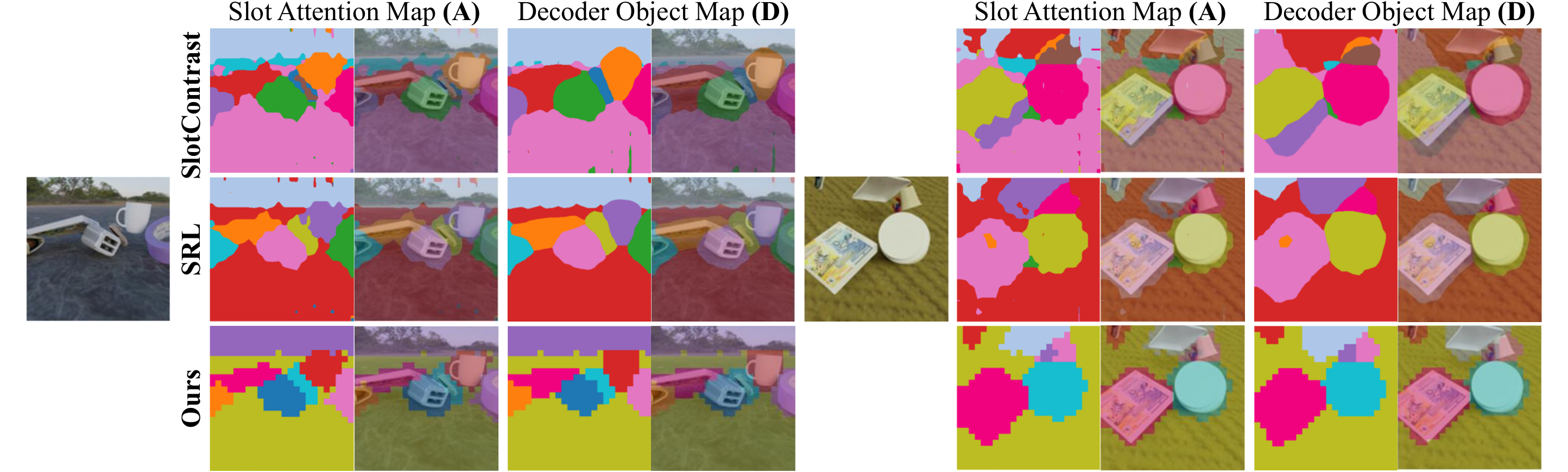} 
\vspace{-0.25cm}
\caption{
\textbf{Visualization of the attention map $\mathbf{A}$ and object map $\mathbf{D}$.}
Compared to SlotContrast~\cite{slotcontrast}, SRL~\cite{srl} reduces noise and produces sharper slot assignments.
However, dense alignment also propagates both noise and blur across the two branches.
Consequently, noisy and spatially blurry representations are still observed in both maps, indicating incomplete resolution of encoder–decoder discrepancy.
}
\label{fig.srlfail}
\renewcommand{\arraystretch}{0.9}
\vspace{-15pt}
\end{figure}
SRL~\cite{srl} attempted to reduce discrepancy between $\mathbf{A}$ and $\mathbf{D}$ via dense alignment~(e.g., contrastive objectives~\cite{supcon, simclr} across the full spatio-temporal volume). 
While straightforward, it implicitly assumes that all patches provide equally reliable supervision; when this assumption fails, indiscriminate agreement propagates erroneous signals from unreliable artifacts.
Consequently, SRL only partially resolves the discrepancy, as each module inadvertently reinforces the other's inherent flaws. 
We provide quantitative and qualitative evidence of this failure in Fig.~\ref{fig.srlfail} and Tab.~\ref{tab:ad_ari}.
Also, dense patch-wise 
\begin{wraptable}{r}{0.4\linewidth}
\vspace{-33pt}
\scriptsize
\renewcommand{\arraystretch}{0.88}  
\setlength{\tabcolsep}{1.2pt} 
\centering
\caption{
\textbf{Encoder-decoder consistency.}
We used symmetric Adjusted Rand Index~(ARI) and asymmetric mean Best Overlap~(mBO).
}
\label{tab:ad_ari}
\begin{tabular}{lccc}
\toprule
\multirow{2}{*}{\textbf{Method}} & ARI & mBO & mBO \\
&$\mathbf{A}\!\leftrightarrow\!\mathbf{D}$ &$\mathbf{A}\!\rightarrow\!\mathbf{D}$ & $\mathbf{D}\!\rightarrow\!\mathbf{A}$ \\
\midrule
SlotContrast & 73.0 & 63.6 & 62.9 \\
SRL & 77.6 & 65.4 & 60.8 \\
Ours & \best{85.7} & \best{72.2} & \best{68.6} \\
\bottomrule
\end{tabular}
\vspace{-24pt}
\end{wraptable}
objective requires pairwise spatio-temporal comparisons, resulting in quadratic complexity $\mathcal{O}((T\cdot H\cdot W)^2)$, which limits its scalability to long or high-resolution videos.

\textbf{Motivation: Selective Alignment Principle.}
Above observations reveal that dense alignment requires simultaneous reliability of both maps, while encoder-decoder asymmetry during optimization often invalidates this assumption.
Therefore, we propose a selective alignment principle: mutual supervision should be applied only where each branch is structurally reliable.
Concretely, we leverage encoder-driven boundaries and decoder-driven interiors to align the counterpart branch, proving that minimal supervision is sufficient for robust alignment once reliable regions are identified.
This prevents error propagation caused by indiscriminate agreement and eliminates the need for dense spatio-temporal pairwise comparisons.

\subsection{Structuring Pseudo-Labels}
\label{sec:pseudo_label}
To instantiate the selective alignment principle, we first construct structured pseudo-labels from the model’s internal spatial maps.
These pseudo-labels serve as teacher signals for cross-module refinement.
For each spatio-temporal patch at $(t,p)$, we derive hard slot assignments from the probabilistic encoder attention map $\mathbf{A}$ and decoder object map $\mathbf{D}$:
\begin{equation}
\hat{s}^A_{t,p} = \arg\max_s \mathbf{A}_{t,p,s},
\qquad
\hat{s}^D_{t,p} = \arg\max_s \mathbf{D}_{t,p,s}.
\end{equation}
These provide candidate targets, whose reliability varies across regions.

\textbf{Local Consistency Analysis.}
To identify structurally reliable regions, we measure local slot consistency
separately on the encoder attention assignments and the decoder object-map assignments.
Let $\mathcal{N}_{\mathrm{sp}}(p)$ denote the spatial 8-neighborhood of patch $p$ on the $H \times W$ patch grid. 
We define the spatio-temporal neighborhood of a patch $(t,p)$ by augmenting
$\mathcal{N}_{\mathrm{sp}}(p)$ with temporally adjacent patches at the same spatial location to further detect motion edges:
\begin{equation}
\mathcal{N}(t,p)
=
\{(t,q)\mid q \in \mathcal{N}_{\mathrm{sp}}(p)\}
\;\cup\;
\{(t-1,p),(t+1,p)\},
\label{Eq:neighbors}
\end{equation}
where out-of-range temporal indices are omitted.
For $\textbf{X}\!\in\!\{\textbf{A}, \textbf{D}\}$, we quantify the local disagreement and agreement counts as:
\begin{equation}
c^{\neq,\textbf{X}}_{(t,p)}
=
\sum_{(t',q)\in \mathcal{N}(t,p)}
\mathbb{I}\big[\hat{s}^\textbf{X}_{t',q} \neq \hat{s}^\textbf{X}_{t,p}\big],
\qquad
c^{=,\textbf{X}}_{(t,p)}
=
\sum_{(t',q)\in \mathcal{N}(t,p)}
\mathbb{I}\big[\hat{s}^\textbf{X}_{t',q} = \hat{s}^\textbf{X}_{t,p}\big].
\label{eq.localneighborcheck}
\end{equation}

\textbf{Boundary Region Selection.}
Boundary regions are characterized by local disagreement while still maintaining semantic grounding.
Since encoder attention assignments are typically sharper, we detect boundary candidates from $\mathbf{A}$:
\begin{equation}
\mathcal{P}_{\mathrm{bd}} = \left\{ (t,p) \mid c^{\neq,\mathbf{A}}_{(t,p)} > n_{\mathrm{bd}} \;\land\; c^{=,\mathbf{A}}_{(t,p)} \geq 1 \right\},
\label{eq:boundary_select}
\end{equation}
where $n_{\mathrm{bd}}$ is a predefined threshold that controls the sensitivity of boundary detection.
This criterion captures meaningful object transitions while excluding isolated noisy assignments by requiring at least one additional agreeing neighbor.

\textbf{Interior Region Selection.}
In contrast, interior~(non-boundary) regions exhibit high local consistency, which is more reliably reflected in the decoder object map assignments.
We thus define the set of interior patches $\mathcal{P}_{\mathrm{nbd}}$ using $\mathbf{D}$:
\begin{equation}
\mathcal{P}_{\mathrm{nbd}} = \left\{ (t,p) \mid c^{\neq,\mathbf{D}}_{(t,p)} < n_{\mathrm{nbd}} \right\},
\label{eq:nonboundary_select}
\end{equation}
where $n_{\mathrm{nbd}}$ controls the strictness of interior selection; a patch is included in $\mathcal{P}_{\mathrm{nbd}}$ only if the number of differing neighbor assignments is strictly smaller than $n_{\mathrm{nbd}}$.
Consequently, these boundary and interior sets~($\mathcal{P}_{\mathrm{bd}}$ and $\mathcal{P}_{\mathrm{nbd}}$) provide reliable regions for
asymmetric cross-distillation: boundary patches supervise the decoder using encoder-derived pseudo-labels, while interior patches supervise the encoder using decoder-derived pseudo-labels.

From a broader perspective, our selection acts as a relaxed morphological erosion~\cite{morphology} parameterized by a $3\times3$ kernel of all ones.
Unlike erosion, which strictly requires all neighboring pixels within a kernel to agree~(making it highly susceptible to noise patches), our parameterized spatiotemporal formulation explicitly filters noise and overcomes standard erosion's lack of temporal awareness.

\subsection{Transitive Pseudo-Label Merging}
\label{sec:merge}
Although selective alignment restricts supervision to structurally reliable regions of each module, we note that pseudo-labels are still imperfect as they are generated on-the-fly from intermediate model predictions.
As a result, they may inherit systematic semantic errors such as over-fragmentation\footnote{Over-fragmentation arises when multiple slots redundantly cooperate to reconstruct a single object in order to minimize reconstruction loss~\cite{adaptiveslot}.}, where a single object is split across multiple slot identities.
When such fragmented assignments are directly used as supervision targets, the labels themselves become inherently flawed. 
This inadvertently reinforces over-fragmentation by forcing the model to associate a single semantic object with multiple, conflicting slot identities.
To prevent this instability, we refine pseudo-labels before applying selective alignment losses.
Specifically, we detect redundant slots based on spatio-temporal activation overlap and consolidate them using a transitive connectivity criterion.

For each slot $s$, we determine whether a spatio-temporal patch is active by thresholding its attention value against the slot-wise mean activation $\mu_s$:
\begin{equation}
\mathbf{M}_{t,p,s} 
=
\mathbb{I}\Big[\mathbf{A}_{t,p,s} > \mu_s\Big],
\qquad
\mu_s = \frac{1}{TP} \sum_{t,p} \mathbf{A}_{t,p,s}.
\label{eq.attnmapanalysis}
\end{equation}
This yields a binary mask $\mathbf{M}$ that identifies regions where each slot exhibits above-average activation over space and time.
Using the binary active-region masks $\mathbf{M}$, we measure the overlap between slots $(s, s')$ via frame-averaged IoU:
\begin{equation}
\label{eq.attnmapiou}
\mathrm{IoU}(s,s') =
\frac{1}{T}\sum_{t=1}^{T}
\frac{\sum_{p}\mathbf{M}_{t,p,s}\mathbf{M}_{t,p,s'}}
{\sum_{p}\mathbf{M}_{t,p,s} + \sum_{p}\mathbf{M}_{t,p,s'} - \sum_{p}\mathbf{M}_{t,p,s}\mathbf{M}_{t,p,s'}}.
\end{equation}
We construct a redundancy graph $\mathcal{G}=(\mathcal{V},\mathcal{E})$,
where each node $s \in \mathcal{V}$ corresponds to a slot identity.
An undirected edge $(s,s') \in \mathcal{E}$ is added if
\begin{equation}
\label{eq.attnmapthr}
\mathrm{IoU}(s,s') > \tau_{\mathrm{merge}},
\end{equation}
indicating that the two slots exhibit substantial spatio-temporal overlap and are therefore likely to represent the same object.

Since redundancy may occur transitively~(e.g., $s_1$ overlaps with $s_2$, and $s_2$ overlaps with $s_3$), we group redundant slots by computing the connected components of $\mathcal{G}$.
Each connected component $\mathcal{C}_k \subset \mathcal{V}$ represents a cluster of mutually redundant slots.
We denote the set of all disjoint clusters as $\mathbb{C} = \{\mathcal{C}_1, \mathcal{C}_2, \dots, \mathcal{C}_m\}$.
For each component $\mathcal{C}_k$, we select a dominant slot identity:
\begin{equation}
s^\star_k =
\arg\max_{s \in \mathcal{C}_k}
\sum_{t,p}\mathbb{I}\big[\hat{s}^A_{t,p}=s\big],
\end{equation}
where the slot with the largest footprint is chosen as the representative identity.

We then define a relabeling function $\phi(\cdot)$ and apply $\phi$ to unify pseudo-label assignments to provide coherent object identities for selective alignment:
\begin{equation}
\phi(s)=
\begin{cases}
s^\star_k & \text{if } s \in \mathcal{C}_k, \\
s & \text{otherwise}.
\end{cases}
\end{equation}
\begin{equation}
\hat{s}^A_{t,p} \leftarrow \phi(\hat{s}^A_{t,p}),
\qquad
\hat{s}^D_{t,p} \leftarrow \phi(\hat{s}^D_{t,p}).
\end{equation}
Using the relabeled pseudo-labels, we recompute the local consistency counts and re-derive the boundary/interior sets $\mathcal{P}_{\mathrm{bd}}$ and $\mathcal{P}_{\mathrm{nbd}}$ via Eq.~(\ref{eq.localneighborcheck})-(\ref{eq:nonboundary_select}).
For simplicity, we reuse $\hat{s}^{A}_{t,p}$ and $\hat{s}^{D}_{t,p}$ to denote the merged pseudo-labels after applying $\phi(\cdot)$.

\subsection{Training Objective}
Given the refined pseudo-labels obtained after transitive merging,
we perform asymmetric cross-distillation between the encoder attention map $\mathbf{A}$ and the decoder object map $\mathbf{D}$.
For boundary patches at $(t,p)\in\mathcal{P}_{\mathrm{bd}}$, the decoder is supervised using one-hot pseudo-labels derived from the encoder:
\begin{equation}
\mathcal{L}_{\mathrm{bd}} =
\frac{1}{|\mathcal{P}_{\mathrm{bd}}|}
\sum_{(t,p)\in\mathcal{P}_{\mathrm{bd}}}
\left\|
\mathbf{D}_{t,p}
-
\mathrm{onehot}(\hat{s}^A_{t,p})
\right\|_2^2,
\label{Eq:Lbd}
\end{equation}
while the interior $(t,p)\in\mathcal{P}_{\mathrm{nbd}}$ of the encoder is supervised from the decoder as:
\begin{equation}
\mathcal{L}_{\mathrm{nbd}} =
\frac{1}{|\mathcal{P}_{\mathrm{nbd}}|}
\sum_{(t,p)\in\mathcal{P}_{\mathrm{nbd}}}
\left\|
\mathbf{A}_{t,p}
-
\mathrm{onehot}(\hat{s}^D_{t,p})
\right\|_2^2.
\label{Eq:Lnbd}
\end{equation}
We adopt an MSE objective rather than cross-entropy, as its scale compatibility with the reconstruction loss enables stable joint optimization without loss reweighting, and its bounded gradients improve robustness to imperfect pseudo-labels during early training~\cite{ghosh2017robust, GCE}.

Finally, the selective alignment losses are combined with the base objective $\mathcal{L}_{\mathrm{base}}$, which includes reconstruction and temporal slot contrastive objective~\cite{slotcontrast} following SRL~\cite{srl}.
Following a warm-up phase covering the first $30\%$ of the total iterations $\eta_{t}$, the complete training objective is defined as:
\begin{equation}
\mathcal{L} = \mathcal{L}_{\mathrm{base}} + \mathbb{I}{[\eta > 0.3\eta_\mathrm{t}]} \cdot \lambda_{\mathrm{SSync}} \left( \mathcal{L}_{\mathrm{bd}} + \mathcal{L}_{\mathrm{nbd}} \right),
\end{equation}
where $\eta$ denotes the current iteration and $\lambda_{\mathrm{SSync}}$ is the coefficient for SSync.

%% file: sec/4_experiment.tex
\section{Experiments}
\subsection{Evaluation Settings}
\textbf{Datasets and Metrics.}
Following prior works~\cite{videosaur, slotcontrast, srl}, we run experiments on three standard VOCL benchmarks: MOVi-C and MOVi-E~\cite{greff2022kubric}, and YouTube-VIS~(YTVIS) 2021~\cite{ytvis,ytvis2021_19,ytvis2021_21}.
The primary challenge of MOVi-C is over-fragmentation due to complex object interactions, whereas MOVi-E contains a large number of small-scale objects, making boundary precision critical.
YTVIS further evaluates robustness under real-world video dynamics.
To further assess generalizability, we also evaluate under the RandSF.Q protocol~\cite{randsf}, benchmarking on lower-resolution variants of MOVi-C and MOVi-D, as well as the High-Quality YTVIS dataset.
Lastly, to verify generalization beyond video domain, we report image-level object-centric performance on MOVi-E and COCO2017~\cite{coco}.

For metrics, we use Foreground ARI~(FG-ARI)~\cite{rand1971objective, fgari} and mBO~\cite{seitzer2023bridging}.
FG-ARI measures permutation-invariant clustering consistency between predicted slot assignments and ground-truth~(GT) masks over foreground pixels.
mBO evaluates object-level coverage by computing the maximum Intersection-over-Union (IoU) between each GT object and predicted slots.
We also report ARI and mIoU when available in prior work.

\input{tab/tab_1_video}
\input{tab/tab_5_randsf}
\input{tab/tab_1_image}
\textbf{Implementation Details.}
We adopt prior training protocols~\cite{slotcontrast, srl, randsf} to ensure fair comparison; all architectural components and base objectives follow the respective baselines.
SSync additionally introduces $n_{\mathrm{bd}}$, $n_{\mathrm{nbd}}$, $\lambda_\mathrm{SSync}$ and $\tau_{\mathrm{merge}}$.
Among them, we fix $n_{\mathrm{bd}}=n_{\mathrm{nbd}}=1$ and $\lambda_{\mathrm{SSync}}=1.0$ across all datasets, which demonstrates SSync's robustness and its minimal requirement for dataset-specific tuning.
We only adjust the merging threshold $\tau_{\mathrm{merge}}$, setting it to $0.7$ for MOVi-C, $0.65$ for MOVi-E, and $0.6$ for YTVIS.
This accounts for varying object densities, reflecting theoretically established principles in clustering where overlap criteria must adapt to spatial density~\cite{adaptivenms, dbscan}.
Importantly, we observe that performance remains consistent within a reasonable range around 2/3, indicating that SSync does not rely on exhaustive threshold tuning.
Details are in the Appendix.

\subsection{Comparison to the State-of-the-art~(SOTA) Methods}

\textbf{Video Benchmarks.}
Comparison to SOTA VOCL methods is shown in Tab.~\ref{tab:vocl}, where SSync consistently achieves superior or highly competitive performance compared to prior approaches across benchmarks.
On MOVi-C, where objects are frequently decomposed into multiple slots, SSync achieves the best results by consolidating fragmented identities through selective alignment.
SSync also yields substantial gains on MOVi-E, where the primary challenge is to accurately capture the boundaries of numerous small objects.
Lastly, on YTVIS dataset, SSync achieves competitive FG-ARI while obtaining the highest mBO, indicating stronger object coverage.
It is worth noting that YTVIS provides GT annotations only for a sparse set of salient foreground objects.
Therefore, these metrics may not fully capture qualitative differences outside the evaluated objects. 
Accordingly, we provide qualitative comparisons in the Appendix; in the shown examples, SSync appears to distinguish secondary objects and background regions more consistently than competing methods. 
Overall, these results suggest that selective alignment more effectively resolves encoder–decoder discrepancy than the dense alignment strategy, despite its simplicity, requiring no additional architectural components used by SRL~\cite{srl}.

In addition, we integrate SSync into two variants of RandSF.Q~\cite{randsf} built upon VideoSAUR~\cite{videosaur} and SlotContrast~\cite{slotcontrast}.
As reported in Tab.~\ref{tab.randsf}, SSync consistently improves performance, which indicates that SSync functions as a plug-and-play method, reinforcing its practical applicability.

\textbf{Image Benchmarks.}
To evaluate generalization beyond video benchmarks, we assess SSync on image-level object-centric benchmarks.
To illustrate, SSync achieves a SOTA FG-ARI of 86.0 on MOVi-E, and attains an ARI of 47.9 and mBO of 33.1 on real-world COCO2017, substantially outperforming SRL.
This suggests that the proposed SSync improves spatial consistency independently of motion signals, further validating its robustness across diverse visual domains.


\begin{wraptable}{r}{4.4cm}
    \vspace{-1cm}
    \centering
    \scriptsize
    \caption{
    \textbf{Memory comparison~(SRL/SSync).} 
    Values represent the maximum VRAM allocated per GPU in GB. 
    OOM indicates Out-Of-Memory errors on an NVIDIA RTX PRO 6000 Blackwell GPU~(97GB). 
    }
    \label{tab:memory}
    \renewcommand{\arraystretch}{0.85} 
    \setlength{\tabcolsep}{1pt}
    \begin{tabular}{l cc}
        \toprule
        \multirow{2}{*}{Frames ($T$)} & \multicolumn{2}{c}{Batch Size per GPU} \\
        \cmidrule(lr){2-3}
        & 32 & 64 \\
        \midrule
        $T=4$ & 70 / \textbf{27} & OOM / \textbf{59} \\
        $T=6$ & OOM / \textbf{48} & OOM / \textbf{89} \\ 
        $T=8$ & OOM / \textbf{60} & OOM / \textbf{93} \\
        \bottomrule
    \end{tabular}
    \vspace{-0.8cm}
\end{wraptable}
\textbf{Memory Efficiency.}
We compare the memory footprint against SRL~\cite{srl} using mixed-precision~(FP16) on YTVIS~(518×518 resolution) in Tab.~\ref{tab:memory}~(maximum VRAM per GPU).
SRL exhibits quadratic memory growth~($\mathcal{O}((T \cdot H \cdot W)^2)$) due to dense patch-to-patch comparisons, whereas SSync scales approximately linearly with the number of patches.
As a result, at batch size 32 per GPU~($T{=}4$), SRL requires 70GB while SSync uses only 27GB, reducing memory usage by roughly 60\%~(and only a marginal 5\% increase over SlotContrast~\cite{slotcontrast}).
Furthermore, SRL already encounters OOM at larger temporal lengths or batch sizes on our GPU, whereas SSync remains feasible.
Note that per GPU VRAM values correspond to the maximum reserved memory, thereby not strictly linear with increasing $T$ due to the properties of CUDA and cuDNN.

\subsection{Ablation Study}
All studies are conducted on MOVi-C.

\textbf{Component Ablation.}
We analyze the contribution of each component in Tab.~\ref{tab:ablation}.
Baseline model achieves 69.0 FG-ARI and 30.6 mBO. Applying boundary supervision~($\mathcal{L}_{\mathrm{bd}}$) alone improves performance to 72.9 FG-ARI and 33.4 mBO.
This confirms that selectively distilling sharp boundary cues from the encoder effectively refines decoder object maps. Similarly, interior supervision~($\mathcal{L}_{\mathrm{nbd}}$) alone
\input{tab/tab_2_ablation}
\!yields 71.4 FG-ARI and 33.3 mBO, indicating that decoder-guided denoising stabilizes noisy encoder assignments, leading to stable training.
When both selective alignment losses are combined, performance increases substantially to 77.1 FG-ARI and 38.0 mBO, demonstrating that boundary calibration and interior denoising are complementary.
Finally, incorporating transitive merging further improves performance to 79.4 FG-ARI and 39.5 mBO.
This additional gain highlights the importance of stabilizing pseudo-label identities before cross-distillation.
Overall, these results validate the selective alignment principle: region-aware supervision improves both boundary precision and slot consistency, while redundancy-aware merging ensures global semantic coherence.

\textbf{Robustness to Number of Slots.}
A major challenge in video slot attention is over-fragmentation, which worsens when the number of slots~($S$) exceeds the actual objects in a scene. 
This forces prior works~\cite{slotcontrast, srl} to carefully tune $S$ per dataset to prevent multiple slots from undesirably cooperating to reconstruct a single object. 
As shown in Tab.~\ref{tab:numslot}, when prior approaches are trained with varying slot counts~($S \in \{7, 11, 15\}$), the performance deteriorates significantly at $S=15$, indicating severe sensitivity to slot over-parameterization.
In contrast, SSync maintains consistent performance across all configurations. 
This stability is driven by our transitive pseudo-label merging, which leverages spatio-temporal consistency to consolidate overlapping, fragmented slots into unified object representations. 
By inherently counteracting over-fragmentation, SSync drastically reduces sensitivity to the hyperparameter $S$, alleviating the burden of dataset-specific tuning and offering a highly practical framework for VOCL.

\begin{table*}[b]
\centering
\scriptsize
\vspace{-10pt}
\caption{
\textbf{Performance comparison with varying number of slots.}
}
\label{tab:numslot}
\vspace{-5pt}
\renewcommand{\arraystretch}{0.89}  
\setlength{\tabcolsep}{8.2pt} 
\begin{tabular}{l cc cc cc}
\toprule
\multirow{2}{*}{\textbf{Method}} &
\multicolumn{2}{c}{\textbf{slot=7}} &
\multicolumn{2}{c}{\textbf{slot=11}} &
\multicolumn{2}{c}{\textbf{slot=15}} \\
\cmidrule(lr){2-3}\cmidrule(lr){4-5}\cmidrule(lr){6-7}
& FG‑ARI\,$\uparrow$ & mBO\,$\uparrow$
& FG‑ARI\,$\uparrow$ & mBO\,$\uparrow$
& FG‑ARI\,$\uparrow$ & mBO\,$\uparrow$ \\ 
\midrule
SlotContrast~\cite{slotcontrast} & 74.9 & 27.9 & 69.3 & 32.7 & 61.8 & \second{31.2} \\
SRL~\cite{srl} & \second{76.5} & \second{31.6} & \second{74.3} & \second{34.5} & \second{72.8} & 31.1 \\
Ours & \best{76.9} & \best{39.8} & \best{79.4} & \best{39.5} & \best{78.8} & \best{41.0} \\
\bottomrule
\end{tabular}
\end{table*}

\textbf{Transitive Pseudo-Label Merging.}
A key design choice in transitive merging is the source of active-region extraction used to identify redundant slots.
While we derive active regions from the encoder attention map $\mathbf{A}$, we evaluate alternative formulations in Tab.~\ref{tab:merging}~(a).
Specifically, we compare: 
(a0) deriving active regions solely from the decoder object map $\mathbf{D}$; 
(a1–a2) hybrid logical criteria, where redundancy is independently computed from both $\mathbf{A}$ and $\mathbf{D}$ via Eq.~\ref{eq.attnmapanalysis}–\ref{eq.attnmapthr} and combined using union~(merge if either suggests redundancy) or intersection~(merge only if both agree); and 
(a3) averaging IoU scores from $\mathbf{A}$ and $\mathbf{D}$~(Eq.~\ref{eq.attnmapiou} applied to each map) prior to thresholding with $\tau_{\mathrm{merge}}$ in Eq.~\ref{eq.attnmapthr}.
Although using $\mathbf{D}$ as the merging source yields slightly lower performance, the overall results remain consistently strong across all configurations.
\input{tab/tab_3_merging}
\!This robustness suggests that redundancy detection reflects intrinsic spatio-temporal activation patterns, rather than dependence on a specific map representation.

We also compare transitive merging with a pairwise baseline~(Tab.~\ref{tab:merging}(b)).
Whereas ours constructs a connectivity graph over redundant slots and merges all slots within a connected component simultaneously, pairwise merging combines only the most similar slot pair whose IoU exceeds $\tau_{\mathrm{merge}}$.
Results suggest that over-fragmented objects are often distributed across multiple slots, requiring global consolidation rather than pairwise agglomeration.

Finally, we compare our merging strategy with the slot regularization method used in SRL~\cite{srl}~(Tab.~\ref{tab:merging}(c)).
Slot regularization mitigates redundancy by penalizing overlapping slot pairs only during a predefined warm-up stage.
In contrast, our method explicitly consolidates redundant slot identities based on spatio-temporal coverage similarity throughout the training.
Empirical results demonstrate that transitive merging provides a more effective remedy for over-fragmentation, while eliminating the need for heuristic scheduling of regularization phases.

\begin{table}[b]
  \centering
  \vspace{-10pt}
  \scriptsize
  \setlength{\tabcolsep}{2.5pt}
  \renewcommand{\arraystretch}{1}
  \caption{
  \textbf{Hyperparameter analysis.}
  The gray row denotes our default configuration.
  }
  \label{Tab.coef}
  \vspace{-0.5cm}
  \begin{subtable}[t]{0.24\linewidth}
    \centering
    \caption{Impact of $\lambda_\mathrm{SSync}$.}
    \label{Tab.lambda}
    \vspace{-3pt}
    \begin{tabular}{l cc}
    \toprule
    $\lambda_{\mathrm{SSync}}$ & FG‑ARI & mBO \\ 
    \midrule
    0.1 & 74.1 & 37.8 \\
    0.5 & 79.1 & 39.0 \\
    \rowcolor{gray!20}
    1.0 & 79.4 & 39.5 \\
    1.2 & 78.9 & 40.2 \\
    \bottomrule
    \end{tabular}
  \end{subtable}\hfill
  \begin{subtable}[t]{0.24\linewidth}
    \centering
    \caption{Impact of $n_\mathrm{bd}$.}
    \label{Tab.bd}
    \vspace{-3pt}
    \begin{tabular}{c cc}
    \toprule
    $n_\mathrm{bd}$ & FG‑ARI & mBO \\  
    \midrule
    \rowcolor{gray!20}
    1 &  79.4 & 39.5 \\
    2 &  78.9 & 39.3 \\
    3 &  78.2 & 39.4 \\
    4 &  76.6 & 36.7 \\
    \bottomrule
    \end{tabular}
  \end{subtable}\hfill
  \begin{subtable}[t]{0.24\linewidth}
    \centering
    \caption{Impact of $n_\mathrm{nbd}$.}
    \label{Tab.nbd}
    \vspace{-3pt}
    \begin{tabular}{l cc}
    \toprule
    $n_\mathrm{nbd}$ & FG‑ARI & mBO \\ 
    \midrule
    \rowcolor{gray!20}
    1 & 79.4 & 39.5 \\
    2 & 78.9 & 41.2 \\
    3 & 78.8 & 40.1 \\
    4 & 76.7 & 36.6\\
    \bottomrule
    \end{tabular}
  \end{subtable}\hfill
    \begin{subtable}[t]{0.24\linewidth}
    \centering
    \caption{Impact of $\tau_\mathrm{merge}$.}
    \label{Tab.taumerge}
    \vspace{-3pt}
    \begin{tabular}{l cc}
    \toprule
    $\tau_{\mathrm{merge}}$ & FG‑ARI & mBO \\ 
    \midrule
    0.65 & 78.3 & 40.1 \\
    \rowcolor{gray!20}
    0.7 & 79.4 & 39.5 \\
    0.75 & 79.1 & 39.5 \\
    0.8 & 78.2 & 39.0 \\
    \bottomrule
    \end{tabular}
  \end{subtable}\hfill
\end{table}

\textbf{Impact of $\lambda_{\mathrm{SSync}}$.}
$\lambda_{\mathrm{SSync}}$ controls the relative weight of the SSync loss against the base objective.
As shown in Tab.~\ref{Tab.lambda}, SSync consistently improves performance over the baseline~(69.0 FG-ARI, 30.6 mBO).
Performance increases steadily as $\lambda_{\mathrm{SSync}}$ grows until $\lambda_{\mathrm{SSync}} = 1.0$, where selective alignment and reconstruction are well balanced.
Beyond this point, further increasing the weight yields diminishing returns, indicating that excessive alignment may over-constrain the representation.
Nevertheless, performance remains stable for slightly larger values~(e.g., $\lambda_{\mathrm{SSync}} = 1.2$), demonstrating robustness to moderate over-weighting.

\textbf{Impact of $n_{\mathrm{bd}}$ and $n_{\mathrm{nbd}}$.}
$n_{\mathrm{bd}}$ and $n_{\mathrm{nbd}}$ determine the sensitivity of boundary and interior region selection.
Larger $n_{\mathrm{bd}}$ imposes stricter boundary criteria, while larger $n_{\mathrm{nbd}}$ relaxes interior selection.
Since interior regions typically occupy the majority in most scenes, we fix $n_{\mathrm{bd}} = n_{\mathrm{nbd}} = 1$ to balance between the complementary supervision signals.
Yet, our empirical results~(Tab.~\ref{Tab.bd} and \ref{Tab.nbd}) confirm that performance remains stable across a reasonable range of values.

\textbf{Impact of $\tau_{\mathrm{merge}}$.}
The merging threshold $\tau_{\mathrm{merge}}$ controls the required spatio-temporal overlap for consolidating redundant slots.
At the extremes, $\tau_{\mathrm{merge}} = 1.0$ disables merging, while $\tau_{\mathrm{merge}} = 0$ collapses all slots into a single identity.
Tab.~\ref{Tab.taumerge} shows that $\tau_{\mathrm{merge}} = 0.7$ performs best on MOVi-C. 
Yet, the performance remains consistently high for thresholds around $2/3$.
This indicates that redundancy consolidation depends primarily on a clear overlap structure rather than precise threshold tuning, further validating the robustness of transitive merging.

\begin{wraptable}{r}{0.48\textwidth}
\vspace{-35pt}
\caption{\textbf{Effects of encoder denoising and decoder deblurring on MOVi-C}.
FCC$_8$ measures mask fragmentation via the average number of connected components per frame, where connectivity is defined using an 8-neighborhood.
GT FCC$_8$ is 6.27.
Match$_{90}$ reports the number of matched GT–slot pairs achieving at least 90\% GT coverage.
For these qualified pairs, we report the outside leakage~(Leak) which indicates the total number of pixels assigned to the matched slot but lying outside the corresponding GT mask, summed over the spatio-temporal volume at the original video resolution~($\times 10^3$ pixels).
}
\label{tab:movic_diag}
\centering
\scriptsize
\setlength{\tabcolsep}{2pt}
\renewcommand{\arraystretch}{0.95}
\begin{tabular}{lccc}
\toprule
\multirow{2}{*}{\textbf{Method}} & Encoder & \multicolumn{2}{c}{Decoder} \\
\cmidrule(lr){2-2}\cmidrule(lr){3-4}
 & FCC$_8\downarrow$& Match$_{90}$$\uparrow$ & Leak $\downarrow$ \\
\midrule
SlotContrast~\cite{slotcontrast} & 33.20 & 639 & 98.95 \\
SRL~\cite{srl}          & 21.03 & 684 & 86.26 \\
SSync (Ours)      & \best{8.79} & \best{702} & \best{72.02} \\
\bottomrule
\end{tabular}
\vspace{-25pt}
\end{wraptable}

\subsection{Analysis}
\textbf{Analysis of the impact of denoising and deblurring.}
To quantify the impact in reducing isolated noise and boundary spillover, we report two diagnostics on MOVi-C in Tab.~\ref{tab:movic_diag}.
First, the frame-averaged connected components~(FCC) measures fragmentation: for each frame, we count the number of connected components in each per-slot binary mask and sum them over slots, then average across frames.
A lower FCC indicates a reduction in isolated noise and over-fragmentation.
SSync significantly suppresses spurious predictions, approaching the GT reference of 6.27.
Second, we measure boundary spillover to assess spatial precision.
We first match GT objects and predicted slots and retain only matched pairs whose GT coverage is at least 90\%~(\textsc{Match}$_{90}$).
For these pairs, we quantify outside leakage as the number of predicted pixels lying outside the GT mask~(lower indicates deblurred boundary). SSync not only yields more qualified matches than SRL, but also reduces the mean outside leakage per video by 16.5\%.
Collectively, these demonstrate that SSync excels at denoising fragmented assignments and sharpening slot coverage at object boundaries.
Full results are in the Appendix.

\begin{wraptable}{r}{0.27\textwidth}
\vspace{-35pt}
\caption{
\textbf{Boundary $F$-score on MOVi-C.}
}
\label{tab:movic_fscore}
\scriptsize
\centering
\setlength{\tabcolsep}{0.5pt}
\renewcommand{\arraystretch}{0.9}
\begin{tabular}{l c}
\toprule
\textbf{Method} & $F$-score $\uparrow$ \\
\midrule
SlotContrast~\cite{slotcontrast} & 0.184 \\
SRL~\cite{srl} & 0.222 \\
SSync~(Ours)    & \best{0.255} \\
\bottomrule
\end{tabular}
\vspace{-18pt}
\end{wraptable}
\textbf{Boundary Analysis.}
While mBO measures region overlap, it may not fully capture boundary sharpness. 
Thus, we evaluate boundary $F$-score~\cite{perazzi2016benchmark} in Tab.~\ref{tab:movic_fscore}. 
As shown, SSync improves $F$-score, supporting our claim that selective alignment leverages the encoder's boundary-sensitive cues to refine object contours.

\begin{figure*}[t]
    \centering
    \vspace{-5pt}
    \includegraphics[width=1.\columnwidth]{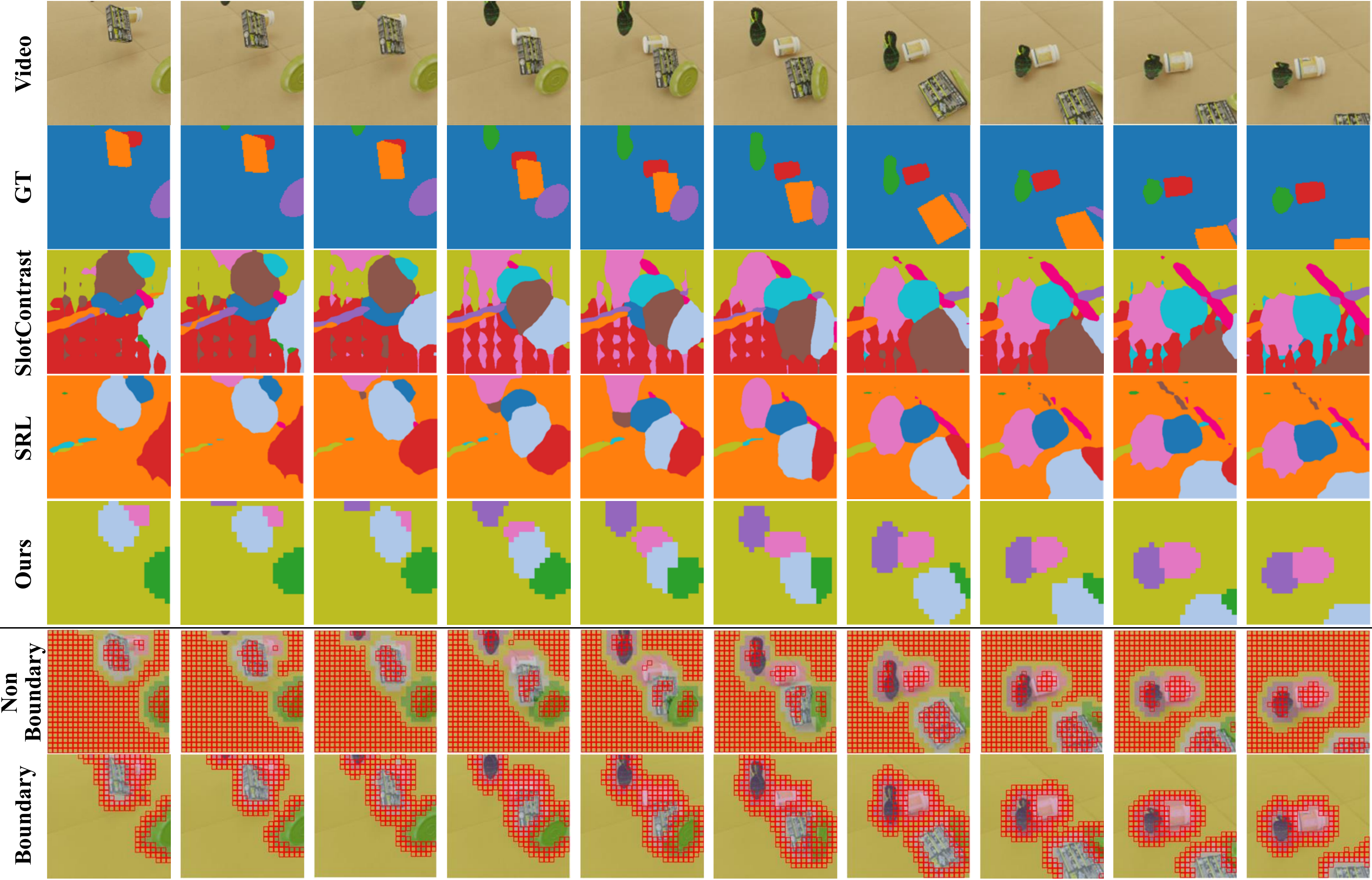} 
    \caption{\textbf{
    Qualitative comparison on MOVi-C.} 
    From top to bottom, we visualize the input, GT masks, predictions from SlotContrast, SRL, and SSync, followed by the non-boundary and boundary regions selected by SSync for interior and boundary supervision, respectively.     
    }
    \label{Fig.movic_main}
    \vspace{-15pt}
\end{figure*}
\textbf{Qualitative Results.}
In Fig.~\ref{Fig.movic_main}, we present qualitative comparisons on MOVi-C.
SlotContrast~\cite{slotcontrast} exhibits severe noisy mask predictions, accompanied by spatially inflated and blurred object boundaries.
Although SRL~\cite{srl} mitigates part of this instability, residual noisy assignments and boundary ambiguity remain evident.
In contrast, SSync produces more coherent object decomposition; boundaries are sharper and spatially consistent, while interior regions exhibit stable semantic grouping.
Notably, background regions~(although not evaluated by foreground-specific metrics like FG-ARI used in Tab.~\ref{tab:vocl}) are consistently represented as unified and temporally coherent entities, while these regions are often fragmented across multiple slots in competing works.
Also, we visualize the selected boundary and interior regions at the bottom of Fig.~\ref{Fig.movic_main}.
Results illustrate that the model effectively separates spatio-temporal object transitions from interior regions across the video sequence.
Additional qualitative comparisons and detailed region-selection visualizations are provided in the Appendix.

%% file: tab/tab_1_video.tex
\begin{table*}[t]
\centering
\vspace{-5pt}
\scriptsize
\caption{\textbf{Experimental results.}
Results are averaged across 3 runs. 
The \hlcaption{bestred}{best} and the \hlcaption{secondorange}{second best} results are denoted by red and orange.
}
\label{tab:vocl}
\vspace{-5pt}
\renewcommand{\arraystretch}{0.93}  
\setlength{\tabcolsep}{3pt} 
\begin{tabular}{lc cc cc cc}
\toprule
\multirow{2}{*}{\textbf{Method}} & \multirow{2}{*}{\textbf{Venue}} & \multicolumn{2}{c}{\textbf{MOVi‑C}} & \multicolumn{2}{c}{\textbf{MOVi‑E}} & \multicolumn{2}{c}{\textbf{YouTube-VIS}} \\
\cmidrule(lr){3-4}\cmidrule(lr){5-6}\cmidrule(lr){7-8}
& & FG‑ARI\,$\uparrow$ & mBO\,$\uparrow$
& FG‑ARI\,$\uparrow$ & mBO\,$\uparrow$
& FG‑ARI\,$\uparrow$ & mBO\,$\uparrow$ \\ 
\midrule
SAVi~\cite{savi} & ICLR'22 & 22.2 & 13.6 & 42.8 & 16.0 & - & - \\
STEVE~\cite{steve} & NeurIPS'22& 36.1 & 26.5 & 50.6 & 26.6 & 15.0 & 19.1 \\
VideoSAUR~\cite{videosaur} & NeurIPS'23 & 64.8 & \second{38.9} & 73.9 & \best{35.6} & 28.9 & 26.3 \\
VideoSAURv2~\cite{videosaur} & NeurIPS'23 & --   & --   & 77.1 & 34.4 & 31.2 & 29.7 \\
SlotContrast~\cite{slotcontrast} & CVPR'25 & 69.3 & 32.7 & 82.9 & 29.2 & 38.0 & 33.7 \\ 
SRL~\cite{srl} & ICLR'26 & 74.3 & 34.5 & 81.9 & 29.3 & \second{42.9} & \second{35.6}  \\ %
SlotCurri~\cite{slotcurri} & CVPR'26 & \second{77.6$_{\pm0.9}$} & 32.8$_{\pm0.2}$ & \second{83.7$_{\pm0.2}$} & 28.9$_{\pm0.7}$ & \best{44.8$_{\pm1.2}$} & 35.5$_{\pm2.2}$  \\ %
SSync~(Ours) & & \best{79.4$_{\pm0.6}$} & \best{39.5$_{\pm0.1}$} & \best{84.0$_{\pm0.9}$} & \second{34.8$_{\pm1.9}$} & 42.6$_{\pm0.2}$ & \best{38.7$_{\pm0.6}$}  \\ %
\bottomrule
\end{tabular}
\vspace{-5pt}
\end{table*}

%% file: tab/tab_5_randsf.tex
\begin{table}[t]
\renewcommand{\arraystretch}{0.92}
\caption{\textbf{Results under the RandSF.Q protocol}~\cite{randsf}. 
$tsim$ denotes the time similarity loss from VideoSAUR~\cite{videosaur}, and SSC refers to the temporal slot contrastive loss from SlotContrast~\cite{slotcontrast}. 
All baselines are reproduced using their official implementations.
}
\label{tab.randsf}
\vspace{-5pt}
\centering
\scriptsize
\setlength{\tabcolsep}{1.3pt}
\begin{tabular}{lcccccccccccc}
\toprule
& \multicolumn{4}{c}{\textbf{MOVi-C}}
& \multicolumn{4}{c}{\textbf{MOVi-D}}
& \multicolumn{4}{c}{\textbf{HQ-YTVIS}} \\
\cmidrule(lr){2-5}\cmidrule(lr){6-9}\cmidrule(lr){10-13}
\textbf{Method} & ARI & FGARI & mBO & mIoU & ARI & FGARI & mBO & mIoU & ARI & FGARI & mBO & mIoU \\
\midrule
RandSF.Q$_{\text{tsim}}$~\cite{randsf}
& 70.7 & 63.3 & 31.1 & 28.1
& 39.3 & 70.5 & 25.6 & 24.3
& 39.2 & 56.3 & 37.2 & 37.0 \\
RandSF.Q$_{\text{tsim}}$+SSync
& \best{73.0} & \best{67.3} & \best{32.3} & \best{29.9}
& \best{45.5} & \best{71.2} & \best{27.6} & \best{25.7}
& \best{42.9} & \best{58.6} & \best{39.4} & \best{39.2} \\
\midrule
RandSF.Q$_{\text{ssc}}$~\cite{randsf}
& 52.7 & 67.8 & 24.2 & 22.1
& 37.3 & 85.8 & \best{28.0} & \best{26.8}
& 40.7 & 57.2 & 38.3 & 37.8 \\
RandSF.Q$_{\text{ssc}}$+SSync
& \best{55.5} & \best{71.4} & \best{25.1} & \best{23.1} 
& \best{39.4} & \best{86.5} & 27.9 & \best{26.8}
& \best{48.6} & \best{57.5} & \best{42.1} & \best{41.9} \\
\bottomrule
\end{tabular}
\vspace{-10pt}
\end{table}

%% file: tab/tab_1_image.tex
\begin{wraptable}{r}{3.5cm}
  \centering
  \scriptsize
  \vspace{-12pt}
  \caption{\textbf{Image object-centric learning.}}
  \label{tab:movie_imgfgari}
  \renewcommand{\arraystretch}{0.87}
  \setlength{\tabcolsep}{0.pt}
  \vspace{-0.16cm}
    \begin{subtable}[t]{1\linewidth}
    \caption{Results on MOVi-E.}
    \label{tab:image_movie}
    \centering
    \begin{tabular}{@{}lc@{}}
        \toprule
        \multicolumn{1}{c}{\textbf{Method}} & FG-ARI $\uparrow$ \\
        \midrule
        VideoSAUR~\cite{videosaur}      & 78.4 \\
        SOLV~\cite{aydemir2023self}           & 80.8 \\
        SlotContrast~\cite{slotcontrast}    & 84.8 \\
        SlotCurri &  \second{84.9} \\
        SSync & \best{86.0} \\
        \bottomrule
    \end{tabular}
    \vspace{-0.05cm}
    \end{subtable}
    \centering
    \renewcommand{\arraystretch}{0.85}
    \setlength{\tabcolsep}{0.5pt}
    \begin{subtable}[t]{1\linewidth}
    \caption{Results on COCO.}
    \label{tab:image_coco}
      \begin{tabular}{lcc}
        \toprule
        \textbf{Method} & FG-ARI & mBO \\
        \midrule
        Baseline & 40.5	& 28.8 \\
        SRL~\cite{srl} & 42.8	& \second{29.4} \\
        SlotCurri~\cite{slotcurri} & \second{43.4}	& 28.9 \\
        SSync & \best{47.9} & \best{33.1} \\
        \bottomrule
    \end{tabular}
    \end{subtable}
    \vspace{-20pt}
\end{wraptable}

%% file: tab/tab_2_ablation.tex
\begin{wraptable}{r!}{5cm}
\renewcommand{\arraystretch}{0.92}
    \centering
    \scriptsize
    \vspace{-12pt}
    \caption{\textbf{Component ablation study.}
    $\mathcal{L}_\mathrm{bd}$, $\mathcal{L}_\mathrm{nbd}$, and T.M. represent calibrating the decoder boundary, denoising the attention map, and transitive pseudo-label merging, respectively.}
    \label{tab:ablation}
    \renewcommand{\arraystretch}{1.}  
    \setlength{\tabcolsep}{2pt} 
    \begin{tabular}{ccccc}
        \toprule
        \multicolumn{3}{c}{Selected Components} & \multicolumn{2}{c}{\textbf{MOVi‑C}} \\ 
        \cmidrule(lr){4-5} 
        $\mathcal{L}_\mathrm{bd}$ & $\mathcal{L}_\mathrm{nbd}$ & T.M. & FG‑ARI & mBO \\ 
        \midrule
         &  &  & 69.0 & 30.6 \\ 
        \checkmark &  &  & 72.9 & 33.4 \\ 
         & \checkmark &  & 71.4 & 33.3 \\ 
        \checkmark & \checkmark &  & \second{77.1} & \second{38.0} \\ 
        \checkmark & \checkmark & \checkmark & \best{79.4} & \best{39.5} \\ 
        \bottomrule
    \end{tabular}
    \vspace{-18pt}
\end{wraptable}

%% file: tab/tab_3_merging.tex
\begin{wraptable}{r}{4.6cm}
    \centering
    \vspace{-10pt}
    \scriptsize
    \caption{\textbf{Variants of transitive merging strategies.} A and D denote attention map and decoder object map, respectively.}
    \label{tab:merging}
    \renewcommand{\arraystretch}{0.85}  
    \setlength{\tabcolsep}{2pt} 
        \begin{tabular}{cccccc}
            \toprule
            & Strategy & FG‑ARI\,$\uparrow$ & mBO\,$\uparrow$ \\ 
            \midrule
            \rowcolor{gray!20}
             & Ours  & 79.4 & 39.5 \\ %
            \multicolumn{4}{c}{Varying criteria for merging} \\
            (a0) & D & 77.5 & 38.7 \\
            (a1) & A $\vee$ D & 78.6 & 39.8 \\ 
            (a2) & A $\wedge$ D & 78.9 & 39.1 \\ 
            (a3) & Avg(A, D) & 79.2 & 40.0  \\ %
            \midrule
            \multicolumn{4}{c}{Merging Range} \\
            (b) & Pairwise & 78.2	& 39.3 \\
            \midrule
            \multicolumn{4}{c}{Comparison to Slot Reg.~\cite{srl}} \\
            (c) & Slot Reg.  & 74.6 & 35.8 \\ %
            \bottomrule
        \end{tabular}
        \vspace{-20pt}
\end{wraptable}

%% file: sec/5_conclusion.tex
\section{Conclusion}
In this paper, we proposed Selective Synergistic Learning~(SSync) for VOCL.
SSync achieves impressive performance gains by both effectively and efficiently mitigating the discrepancy between the encoder’s slot attention maps and the decoder’s object maps. 
Specifically, SSync performs selective alignment between the attention map and object map only on regions where each map possesses its respective expertise, enabled by an efficient pseudo-labeling scheme.
To further reduce the risk of pseudo-label corruption caused by object over-fragmentation, we introduced a transitive pseudo-label merging mechanism. 
By analyzing spatio-temporal slot activations and merging redundant slots via a connectivity-based criterion, we refine pseudo-label targets to be more semantically coherent and robust throughout training. 
Extensive experiments across VOCL benchmarks and additional evaluation protocols demonstrate the effectiveness of SSync, while remaining modular and easy to integrate into existing slot-based pipelines.

%% file: sec/appendix.tex
\section{Training Details}
\begin{table}[t]
\scriptsize
\centering
\caption{
\textbf{Hyperparameters for SSync training on MOVi-C, MOVi-E, and YouTube-VIS 2021.}
}
\label{tab:hyperparams}
\resizebox{\textwidth}{!}{%
\begin{tabular}{lccc}
\toprule
\textbf{Hyperparameter} & \textbf{MOVi-C} & \textbf{MOVi-E} & \textbf{YouTube-VIS}  \\ \midrule
\textit{Training Configurations} & & & \\
Training Steps & 100k & 300k & 100k \\
Batch Size & 128 & 128 & 128 \\
Optimizer & Adam & Adam & Adam \\
Learning Rate & 0.0008 & 0.0008 & 0.0008 \\
ViT Architecture & DINOv2-Small & DINOv2-Base & DINOv2-Base \\
Image Size & $336 \times 336$ & $336 \times 336$ & $518 \times 518$ \\ \midrule
\textit{Slot Attention \& Architecture} & & \\
Number of Slots & 11 & 15 & 7 \\
Slot Dimension ($D_{\text{slots}}$) & 64 & 128 & 64 \\
Iterations (first / other frames) & 3 / 2 & 3 / 2 & 3 / 2 \\
Decoder Type & MLP & MLP & MLP \\ \midrule
\textit{SSync Parameters (Ours)} & & & \\
Boundary Sensitivity ($n_{\mathrm{bd}}$) & 1 & 1 & 1 \\
Non-boundary Sensitivity ($n_{\mathrm{nbd}}$) & 1 & 1 & 1 \\
Loss Coefficient ($\lambda_{\text{SSync}}$) & 1.0 & 1.0 & 1.0 \\
Merging Threshold ($\tau_{\mathrm{merge}}$) & 0.7 & 0.65 & 0.6 \\ \bottomrule
\end{tabular}%
}
\end{table}

Regarding the training configurations, we followed SlotContrast~\cite{slotcontrast}\footnote{https://github.com/martius-lab/slotcontrast} and SRL~\cite{srl}\footnote{https://github.com/hynnsk/SRL} to set up the experiments in Tab.~\ref{tab:vocl} and Tab.~\ref{tab:movie_imgfgari}.
Across all datasets, we maintain a uniform batch size of 128 and a learning rate of 8e-4. 
For our hardware resources, all experiments were conducted using 2$\times$ NVIDIA RTX A6000 GPUs. The only exception is the YouTube-VIS dataset; due to its higher input resolution of 518$\times$518, requiring greater VRAM, we performed those specific experiments on the NVIDIA RTX PRO 6000 Blackwell GPUs.

In addition, for Tab.~\ref{tab.randsf}, we followed RandSF.Q~\cite{randsf}.
We use DINOv2-Small for all datasets to process $224\times224$ images, and trained for 50000 iterations for all experiments.
Our hyperparameters~($n_{\mathrm{bd}}$, $n_{\mathrm{nbd}}$, $\lambda_{\text{SSync}}$, and $\tau_{\mathrm{merge}}$) are kept consistent across all benchmarks to demonstrate the robustness of our framework, while we utilize the same value for MOVi-D as established for MOVi-C.
For more details of this setting, we refer to the scripts in the official repository\footnote{https://github.com/Genera1Z/RandSF.Q}.

\begin{table}[t]
\centering
\scriptsize
\caption{\textbf{Detailed Comparison with SRL.} Compared to SRL, SSync eliminates auxiliary projectors and complex contrastive objectives, achieving linear complexity and higher parameter efficiency.}
\label{tab:srl_comparison}
\resizebox{\linewidth}{!}{%
\begin{tabular}{lcc}
\toprule
\textbf{Aspect} & \textbf{SRL~\cite{srl}} & \textbf{SSync (Ours)} \\ \midrule
Alignment Scope & Dense~(all patches) & Selective~(boundary/interior) \\
Alignment Method & Ternary contrastive loss & Pseudo-label MSE \\
Redundancy Handling & Warm-up Reg.~(fixed) & Transitive Merging~(adaptive) \\
Complexity & $\mathcal{O}((T \cdot H \cdot W)^2)$ & $\mathcal{O}(T \cdot H \cdot W)$ \\
Extra Modules & 2 Projectors & None \\ \bottomrule
\end{tabular}%
}
\end{table}

\section{Detailed Comparison with SRL}
\label{sec:comparison_srl}
To further demonstrate the practical advantages of SSync, we provide a detailed comparison with SRL~\cite{srl}, summarized in Table~\ref{tab:srl_comparison}. 
Our design emphasizes simplifying mutual learning while improving scalability, efficiency, and robustness.
Unlike SRL, which introduces two additional MLP projectors to map encoder and decoder features into the embedding space for contrastive learning, SSync operates directly on the native output space~(i.e., slot attention maps). 
This projector-free design introduces no additional parameters, making SSync a fully plug-and-play module that can be integrated into existing slot-based architectures with minimal implementation overhead.

SSync also significantly reduces computational complexity. 
SRL relies on a ternary contrastive loss that computes dense similarities across all spatio-temporal patches, resulting in quadratic complexity, $\mathcal{O}((T \cdot H \cdot W)^2)$. 
In contrast, SSync applies a selective MSE objective only to reliable regions, reducing the complexity to linear time, $\mathcal{O}(T \cdot H \cdot W)$. 
This enables scalable training on higher-resolution inputs and longer video sequences where SRL becomes memory-intensive.

Regarding redundancy mitigation, SRL relies on a slot regularization technique ($\lambda_{\mathrm{reg}}$) paired with a complex warm-up schedule. 
This approach necessitates intricate hyperparameter tuning to synchronize multiple warm-up phases and strictly enforces a uniform distribution to keep exactly half of the slots~(fixed) vacant during the initial stage. 
In contrast, SSync introduces transitive pseudo-label merging, an adaptive strategy that dynamically resolves redundancy throughout the entire training process by responding to the real-time evolution of the scene composition.

Finally, SRL’s ternary contrastive objective enforces dense alignment across all regions, implicitly assuming uniform reliability between encoder and decoder outputs. 
In contrast, SSync reformulates alignment as selective cross-distillation, allowing each branch to supervise only the regions where it is most reliable. 
This targeted supervision avoids propagating structural weaknesses such as encoder noise and decoder blur, leading to improved object decomposition quality.

\begin{figure*}[t]
\centering
\vspace{0pt}
\includegraphics[width=1\columnwidth]{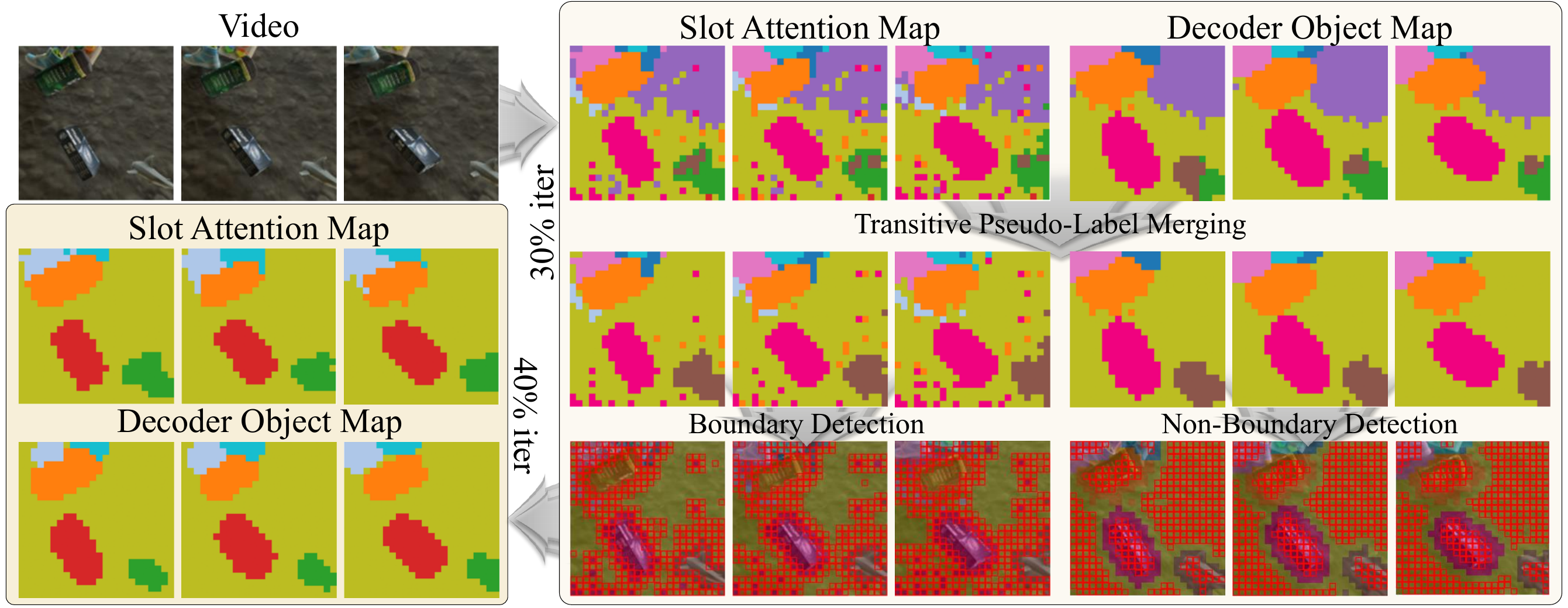} 
\caption{\textbf{Visualization of the Evolution of SSync.} 
}
\label{Fig.ssyncmechanism}
\vspace{-5pt}
\end{figure*}

\section{Qualitative illustration of SSync Mechanisms.}
To provide a comprehensive understanding of SSync, we illustrate the training mechanism of SSync in Fig.~\ref{Fig.ssyncmechanism}.
Our synergistic objectives are activated following the warm-up covering 30\% of the total training iteration. 
At the onset of activation, the attention map exhibits noisy patches, while the object map suffers from blurry object boundaries; also, both suffer from object over-fragmentation.
To extract reliable supervision from these imperfect maps, we first apply transitive merging, which identifies and fuses redundant slots by analyzing their spatio-temporal overlap, yielding semantically coherent groupings even before global convergence. 
The bottom right panels then depict our selective alignment principle: we identify boundary regions~($\mathcal{P}_{bd}$) from the sharp attention maps and non-boundary~(interior) regions~($\mathcal{P}_{nbd}$) from the consistent decoder maps. 
With this reliable mutual supervision, the model refines its prediction iteratively. 
Consequently, as training progresses to 40\% of the total training iterations, SSync results in significantly cleaner object discovery, where each slot captures a unified object identity with precise boundaries and consistent interior semantics.

\begin{figure*}[b]
    \centering
    \includegraphics[width=1.0\columnwidth]{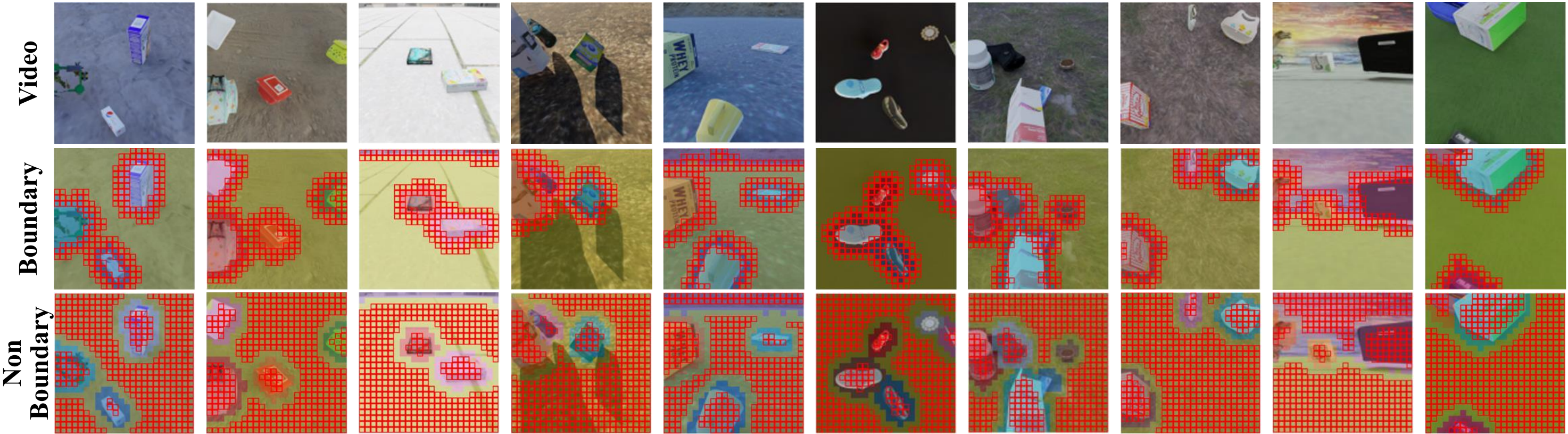} 
    \caption{\textbf{Visualizations of detected boundary and non-boundary patches.} 
    }
    \label{Fig.boundaryanalysis}
\end{figure*}
\section{Boundary \& Non-Boundary (Interior) Analysis}
In this section, we further visualize the patches classified as boundary and interior regions by SSync.
Fig.~\ref{Fig.boundaryanalysis} presents representative frames from 10 example videos.
The visualizations show that SSync effectively identifies object boundary regions and semantically consistent interior regions.
Boundary patches are predominantly located along object transitions, while interior patches concentrate within stable object cores.
This demonstrates that the local consistency criterion successfully captures high-frequency structural transitions without disrupting interior coherence.

We note that, in some cases, patches inside an object are classified as boundary regions.
This occurs when the semantic content at a spatial location changes across consecutive frames due to object motion.
In such cases, the patch is treated as a temporal boundary, allowing the model to capture dynamic transitions along the temporal axis.
This behavior indicates that the boundary selection mechanism is not limited to spatial edges but also adapts to spatio-temporal changes, facilitating improved modeling of object dynamics.

\section{Impact of Denoising and Deblurring}
\begin{table}[t]
\caption{
\textbf{Analysis of encoder denoising via spatial fragmentation on MOVi-C. }
We measure fragmentation using Frame-averaged Total Connected Components~(FCC), where we count the number of connected components for each slot's binary mask per frame and sum them across all slots. A lower FCC indicates a reduction in isolated noise and over-fragmentation.
GT FCC is constant across methods.}
\label{tab:movic_fcc_full}
\centering
\scriptsize
\setlength{\tabcolsep}{6pt}
\renewcommand{\arraystretch}{1.05}
\begin{tabular}{lcccc}
\toprule
& \multicolumn{2}{c}{FCC$_4$} & \multicolumn{2}{c}{FCC$_8$} \\
\cmidrule(lr){2-3}\cmidrule(lr){4-5}
Method & GT & Pred $\downarrow$ & GT & Pred $\downarrow$ \\
\midrule
SlotContrast~\cite{slotcontrast} & 7.273 & 33.785 & 6.268 & 33.205 \\
SRL~\cite{srl}          & 7.273 & \second{21.338} & 6.268 & \second{21.030} \\
SSync~(Ours)      & 7.273 & \best{8.899} & 6.268 & \best{8.793} \\
\bottomrule
\end{tabular}
\end{table}
We report the full analysis results corresponding to Tab.~\ref{tab:movic_diag} to quantify the effects of denoising and deblurring.
First, to validate the denoising effect of SSync, we measure the reduction in the number of fragmented components.
Tab.~\ref{tab:movic_fcc_full} reports frame-averaged connected components~(FCC), which measure spatial fragmentation of predicted slot masks.
SlotContrast exhibits severe over-fragmentation, with FCC values exceeding 33 under both 4- and 8-neighbor connectivity.
SRL partially mitigates this issue, reducing FCC to approximately 21.
In contrast, SSync substantially lowers fragmentation~(FCC$_8$: 8.793), approaching the ground-truth structure~(6.268).
This pronounced reduction indicates that selective interior denoising effectively suppresses noisy slot assignments and consolidates object identities.

\begin{table}[t]
\caption{
\textbf{Analysis of decoder deblurring via measuring outside leakage on MOVi-C.}
For each ground-truth~(GT) object, we match the best corresponding predicted slot and compute the GT coverage of that slot.
We report the number of matched slots achieving at least 75\% and 90\% GT coverage, denoted as Match$_{75}$ and Match$_{90}$, respectively.
For these matched slots, we measure outside leakage, defined as the number of pixels assigned to the slot but lying outside the corresponding GT mask, aggregated over the full spatio-temporal volume of the original video resolution~($24 \times 336 \times 336$).
}
\label{tab:movic_inside_full}
\centering
\scriptsize
\setlength{\tabcolsep}{4pt}
\renewcommand{\arraystretch}{1.05}
\begin{tabular}{lcc|cc}
\toprule
& \multicolumn{2}{c}{GT Coverage $\ge$75\%} & \multicolumn{2}{c}{GT Coverage $\ge$90\%} \\
\cmidrule(lr){2-3}\cmidrule(lr){4-5}
Method & Leak~($\times 10^3$)$\downarrow$ & Match$_\mathrm{75}\uparrow$ & Leak~($\times 10^3)\downarrow$ & Match$_\mathrm{90}\uparrow$ \\
\midrule
SlotContrast~\cite{slotcontrast} & 95.22 & 915 & 98.95 & 639 \\
SRL~\cite{srl}          & \second{82.06} & \second{952} & \second{86.26} & \second{684} \\
SSync~(Ours)      & \best{71.03} & \best{957} & \best{72.02} & \best{702} \\
\bottomrule
\end{tabular}
\end{table}
To further validate the effectiveness of SSync in refining object boundaries, we evaluate outside leakage in Tab.~\ref{tab:movic_inside_full}.
Given a GT-slot pair where the predicted slot mask covers over 75\% or 90\% of the GT mask, we quantify the area of the predicted slot that spills over into the background.
As reported, SSync consistently achieves the lowest leakage across both coverage thresholds while simultaneously increasing the number of high-coverage matches.
These results demonstrate the effectiveness of selectively refining the decoder boundary representation by leveraging the attention maps' boundary sensitivity.
Altogether, the fragmentation and leakage analyses confirm that SSync excels at denoising the representations in the encoder attention map and deblurring the spatially inflated boundaries in the decoder object map by selectively distilling the complementary strengths of each module.

\section{Evolution of Synergistic Refinement}
To further investigate the mutual refinement process between the encoder and decoder, we analyze the evolution of three diagnostic metrics throughout the training: (1) outside leakage~(for boundary blurring), (2) FCC~(for spatial denoising), and (3) the overlap ratio between the encoder-derived boundary set $\mathcal{P}_{\mathrm{bd}}$ and the decoder-derived non-boundary set $\mathcal{P}_{\mathrm{nbd}}$.
We adopt these specific proxies rather than direct region-wise~(boundary and non-boundary) evaluations because the disproportionate scale of interior regions tends to dilute the sensitivity of noise measurements, making direct interior-based metrics less informative.
As shown in Tab.~\ref{tab:evolution_metrics}, the synergistic effect of SSync enables each module to iteratively overcome its inherent inductive biases by leveraging the other's strengths.

\begin{table}[h]
\centering
\scriptsize
\caption{
\textbf{Evolution of deblurring and denoising metrics on MOVi-C.}
We measure boundary blurring via outside leakage~($\times 10^3$ pixels; lower is better) and spatial noise via FCC~(lower is better). 
The synergistic refinement allows the decoder to sharpen boundaries and the encoder to suppress noise over training iterations.
The key indicators of deblurring and denoising are emphasized with \hlcaption{thirdyellow}{yellow} for better clarity.
}
\label{tab:evolution_metrics}
\begin{tabular}{ccccc}
\toprule
Metric & Module & 30\% Iter & 50\% Iter & 100\% Iter \\ \midrule
\rowcolor{gray!20}
\multicolumn{5}{c}{Decoder Deblurring Effect} \\
\multirow{2}{*}{Outside Leakage} \multirow{2}{*}{(GT Coverage $\ge$75\%) $\downarrow$} & Decoder ($\textbf{D}$) & \third{93.96} &	\third{75.20} & \third{71.03} \\
  & Encoder~($\textbf{A}$) & 74.60 & 73.54 & 70.18 \\ 
\multirow{2}{*}{Outside Leakage} \multirow{2}{*}{(GT Coverage $\ge$90\%) $\downarrow$} & Decoder ($\textbf{D}$) & \third{99.21} & \third{80.72} & \third{72.02} \\
  & Encoder~($\textbf{A}$) & 76.89 & 77.89 & 73.65 \\ \midrule
\rowcolor{gray!20}
\multicolumn{5}{c}{Encoder Denoising Effect} \\
\multirow{2}{*}{FCC} \multirow{2}{*}{(4 spatial neighborhood) $\downarrow$} & Decoder ($\textbf{D}$) & 13.75 & 7.01 & 6.88 \\
  & Encoder ($\textbf{A}$) & \third{41.48} & \third{9.39} & \third{8.90} \\ 
\multirow{2}{*}{FCC} \multirow{2}{*}{(8 spatial neighborhood) $\downarrow$} & Decoder ($\textbf{D}$) & 13.15 & 6.89 & 6.76 \\
  & Encoder ($\textbf{A}$) & \third{40.75} & \third{9.29} & \third{8.79} \\ \bottomrule
\end{tabular}
\end{table}
To illustrate, the decoder object map~(\textbf{D}) suffers from blurry boundaries~(high outside leakage) due to the smoothing nature of the reconstruction objective. 
For instance, at 30\% of training, the decoder's outside leakage~(GT Coverage $\ge$75\%) is approximately $93.96 \times 10^3$ patches. 
However, as the decoder receives sharp boundary guidance from the encoder's attention maps through our selective alignment loss~($\mathcal{L}_{bd}$), its leakage significantly drops to $71.03 \times 10^3$ by the end of training. 
This sharp reduction demonstrates that the encoder's boundary sensitivity effectively deblurs the decoder's spatial assignments.

Conversely, the encoder's attention assignments are initially susceptible to noisy predictions, as reflected by a high FCC of $41.48$~(connectivity measured with 4 spatial neighborhood patches) at 30\% iterations. 
Yet, as the inherently less noisy decoder output~(FCC $= 13.15$ at 30\% of total iterations) is distilled to the interior regions of the encoder attention map via $\mathcal{L}_{nbd}$, the encoder's FCC sharply decreases to $8.90$, effectively removing isolated noisy patches and leading to more coherent object discovery.
Collectively, these metrics confirm that the selective synergistic mechanism transforms a potential mismatch between the two modules into mutual improvement.

\begin{table}[t]
\centering
\scriptsize
\caption{
\textbf{Evolution of the overlap ratio between encoder-derived boundary patches
$\mathcal{P}_{bd}$ and decoder-derived non-boundary patches $\mathcal{P}_{nbd}$. }
}
\begin{tabular}{lccc}
\toprule
Metric & 30\% Iter & 50\% Iter & 100\% Iter \\
\midrule
Overlap ratio~(IoU) between $\mathcal{P}_{bd}$ and $\mathcal{P}_{nbd}$ 
& \third{0.277} & \third{0.064} & \third{0.059} \\
\bottomrule
\end{tabular}
\label{tab:evolution_overlap}
\end{table}

Beyond the improvements in leakage and fragmentation, we track the spatial evolution of the two selective supervision sets during training. 
Specifically, we measure the overlap ratio between the encoder-derived boundary patches $\mathcal{P}_{bd}$~(Eq.~\ref{eq:boundary_select}) and the decoder-derived non-boundary patches $\mathcal{P}_{nbd}$~(Eq.~\ref{eq:nonboundary_select}). 
At the activation point~(30\% of iterations), the overlap~(measured with IoU) is relatively high~(0.277), reflecting that the boundary-interior separation remains imperfect immediately following the warm-up phase. 
As training proceeds, the overlap ratio rapidly decreases to 0.064 at 50\% and stabilizes at 0.059 at convergence, indicating that the two supervision sets become increasingly disjoint. 
This trend suggests that the model effectively learns to separate $\mathcal{L}_{bd}$ and $\mathcal{L}_{nbd}$ on distinct spatial regions.

\section{Quality of Learned Slots: Object Dynamics Prediction}
To validate the usefulness of our learned slots, we evaluate their transferability to a downstream task: object dynamics prediction. 
Following prior work, we employ the SlotFormer~\cite{slotformer} framework as the dynamics prediction module on top of our frozen object-centric model to autoregressively predict future slots for $F$ rollout steps, given $B$ burn-in frames. 
Specifically, we set $(B, F)$ to $(14, 10)$, $(5, 10)$, and $(10, 5)$ for the MOVi-C, MOVi-E, and YouTube-VIS datasets, respectively. 
The module is trained for 100,000 iterations with a batch size of 128.
As shown in Tab.~\ref{tab:supple_object_dynamics}, SSync achieves consistent improvements across all datasets compared to previous methods. 
Notably, while prior models struggle with dynamics prediction on MOVi-E due to the prevalence of small objects, SSync excels in modeling fine-grained dynamics by capturing precise semantic boundaries. 
These results demonstrate that SSync not only enhances static object discovery but also produces robust representations that better capture object dynamics in realistic video settings.

\begin{table}[t]
\centering
\scriptsize
\caption{\textbf{Results on object dynamics prediction.}
SlotFormer~(SF) is employed to evaluate the learned slots from each method.
}
\label{tab:supple_object_dynamics}
\renewcommand{\arraystretch}{1.}  
\setlength{\tabcolsep}{6pt} 
\begin{tabular}{l cc cc cc}
\toprule
\multirow{2}{*}{\textbf{Method}} &
\multicolumn{2}{c}{\textbf{MOVi‑C}} &
\multicolumn{2}{c}{\textbf{MOVi‑E}} & \multicolumn{2}{c}{\textbf{YouTube-VIS}} \\
\cmidrule(lr){2-3}\cmidrule(lr){4-5}\cmidrule(lr){6-7}
& FG‑ARI\,$\uparrow$ & mBO\,$\uparrow$
& FG‑ARI\,$\uparrow$ & mBO\,$\uparrow$
& FG‑ARI\,$\uparrow$ & mBO\,$\uparrow$ \\ 
\midrule
Reconstruction + SF & 50.7 & 25.9 & \second{70.6} & 24.3 & 27.4 & 28.9 \\ 
SlotContrast + SF & 63.8 & 26.1 & 70.5 & \second{24.9} & 29.2 & 29.6 \\
SRL + SF & \second{68.9} & \second{27.4} & 70.4 & 24.9 & \best{32.2} & \second{30.0} \\ %
SSync~(Ours) + SF & \best{69.1} & \best{29.0} & \best{72.1} & \best{27.1} & \second{32.1} & \best{30.9} \\ %
\bottomrule
\end{tabular}
\end{table}

\section{Ablation on the SSync Warmup Ratio}
\label{sec:appendix_warmup}
In this section, we investigate the impact of the SSync activation schedule on the final decomposition performance.
By default, SSync activates the selective alignment loss after the first 30\% of training iterations.
To validate the robustness against different warmup schedules, we evaluate the model's performance by varying the warmup length from 5\% to 50\% of the total training steps.
As summarized in Tab.~\ref{tab:warmup_ablation}, the results demonstrate that SSync is generally robust to the choice of warmup length, consistently maintaining competitive scores across different configurations.

\begin{table}[t]
    \centering
    \scriptsize
    \caption{
    \textbf{Performance comparison across varying warmup steps.} 
    }
    \label{tab:warmup_ablation}
    \renewcommand{\arraystretch}{1.1} 
    \setlength{\tabcolsep}{8pt}
    \begin{tabular}{c cc}
        \toprule
        Ratio of Total Iterations & FG-ARI~($\uparrow$) & mBO~($\uparrow$) \\
        \midrule
        5\%  & 79.1 & 40.8 \\
        10\% & 78.6 & 38.3 \\
        15\% & 79.0 & 40.5 \\
        20\% & 78.6 & 40.2 \\
        \rowcolor{gray!15} 30\%~(Default) & 79.4 & 39.5 \\
        50\% & 79.1 & 39.8 \\
        \bottomrule
    \end{tabular}
\end{table}

\begin{table}[t]
    \centering
    \scriptsize
    \caption{
    \textbf{Ablation on transitive merging thresholds.} 
    We replace the default slot-wise mean activation threshold in Eq.~(\ref{eq.attnmapanalysis}) using a quantile-based thresholding strategy. 
    }
    \label{tab:quantile_ablation}
    \renewcommand{\arraystretch}{1.1} 
    \setlength{\tabcolsep}{12pt}
    \begin{tabular}{c cc}
        \toprule
        Threshold Strategy & FG-ARI~($\uparrow$) & mBO~($\uparrow$) \\
        \midrule 
        Ours & 79.4 & 39.5 \\
        \midrule
        Quantile Top 10\% & 79.6 & 37.8 \\
        Quantile Top 20\% & 79.8 & 39.6 \\
        Quantile Top 30\% & 80.2 & 40.4 \\
        \bottomrule
    \end{tabular}
\end{table}
\section{Ablation on Activation Thresholding for Transitive Merging}
\label{sec:appendix_merging_threshold}
In our default SSync framework, the transitive merging module determines whether a spatio-temporal patch is active by thresholding its attention value against the slot-wise mean activation.
To further investigate the sensitivity and potential optimization of this activation criterion, we conduct an ablation study replacing the mean-based threshold with a quantile-based strategy. 
Specifically, instead of the mean-based threshold, we retain only the top $k$-th quantile~(i.e., the top 10\%, 20\%, and 30\%) of the highest activation values per slot to define the active regions for constructing the overlap graph.

As shown in Tab.~\ref{tab:quantile_ablation}, varying the quantile threshold slightly enhances the performance.
The results indicate that tuning the quantile hyperparameter may yield the best overall performance.
However, we point out that employing a fixed quantile introduces a dataset-dependent hyperparameter; since the optimal quantile is inherently tied to the average scale and scale variance of objects within a specific dataset, a fixed threshold may not generalize well to other domains where object sizes significantly differ. 
In contrast, the slot-wise mean activation dynamically adapts to the spatial extent of each object on the fly, providing a parameter-free and highly generalizable criterion. 
Therefore, we adopt the mean activation as our robust default to ensure stable performance across diverse environments without the burden of dataset-specific tuning, while noting that quantile thresholding remains a viable option for domain-specific performance maximization.

\input{tab/tab_7_selection}
\section{Ablation on Patch Selection Variants}
In this section, we compare our boundary and interior selection mechanism with alternative approaches, as detailed in Tab.~\ref{tab:selection_alternative}. 
To isolate the benefits of the selection process, these comparisons are conducted without applying the transitive merging method.

As discussed in Sec.~\ref{sec:pseudo_label}, our selection mechanism essentially operates as a relaxed morphological erosion, but with two key enhancements. 
Since standard erosion is susceptible to noise and ignores temporal dynamics, we incorporate a noise filtering term~(Eq.~\ref{eq:boundary_select}) and extend the spatial comparison to the spatiotemporal domain, enabling the detection of motion edges for temporal coherence. 
Consequently, while standard erosion yields reasonable performance~(validating our core philosophy), our enhanced selection method achieves more significant gains.

Additionally, we evaluate an entropy-based variant that utilizes prediction entropy as a reliability measure. In this setup, we treat low-entropy patches as reliable and selectively use them for distillation. 
As demonstrated, even when varying the entropy threshold from the average value to different percentiles, our straightforward design consistently outperforms these entropy-based alternatives.

\input{tab/tab_8_additional}
\section{Additional Ablation Studies on SSync Design}
Finally, we compare our proposed SSync with alternative designs in Tab.~\ref{tab:other_alternative}. 
First, we implement two soft variants of SSync: one that replaces hard pseudo-labels with soft supervision targets based on the model's internal confidence~(Eq.~(\ref{Eq:Lbd})-(\ref{Eq:Lnbd})), and another that further relaxes the region selection by computing disagreement scores using the raw probability distributions of neighboring patches~(Eq.~(\ref{Eq:neighbors})-(\ref{eq.localneighborcheck})). 
Next, we evaluate a reverse SSync, which extracts interior regions from the encoder attention map and boundary regions from the decoder object map for synergistic learning. 
As shown, our default formulation outperforms all alternatives. 
This shows that using soft distributions as targets inherently retains uncertainty, which can propagate structural ambiguity between modules, while the inferior performance of the reverse strategy explicitly validates our claim that the encoder excels at capturing sharp boundaries, whereas the decoder is better suited for maintaining consistent interiors.

\section{Additional Qualitative Results.}
\begin{figure*}[t]
\centering
\begin{minipage}{1.0\textwidth}
    \centering
    \includegraphics[width=1.0\linewidth]{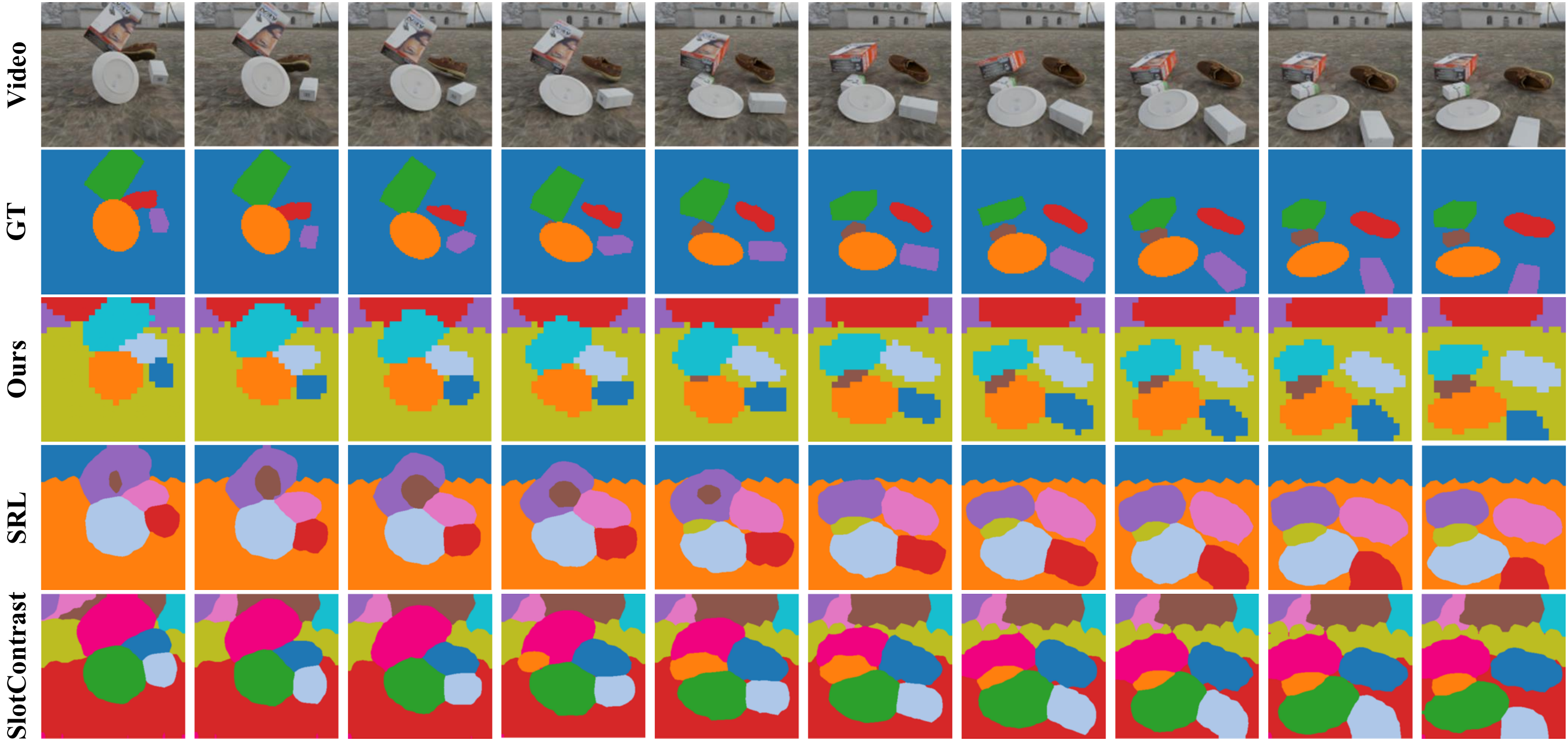}
    \subcaption{Video 1}
    \label{movic_1}
\end{minipage}
\vspace{1em} 
\begin{minipage}{1.0\textwidth}
    \centering
    \includegraphics[width=1.0\linewidth]{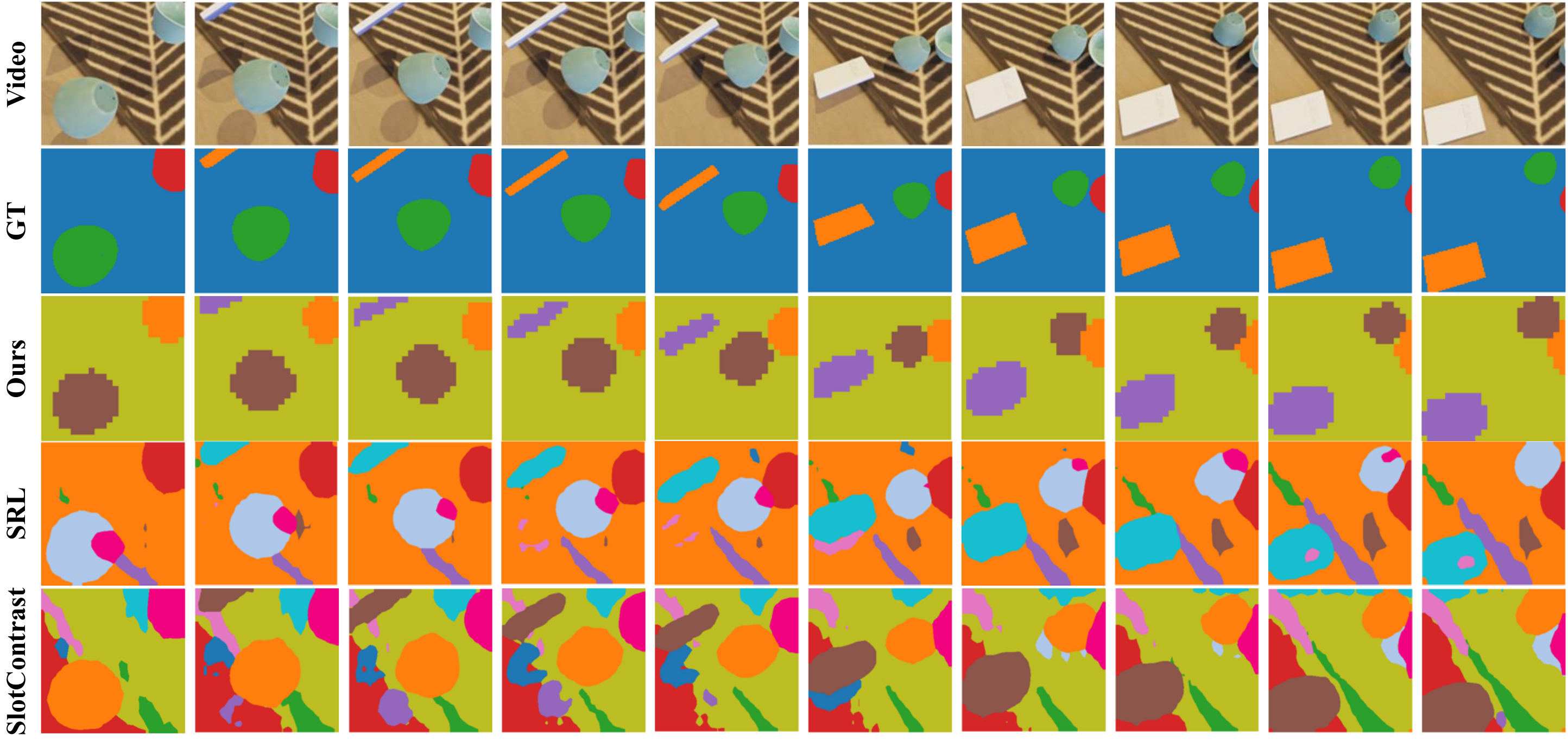}
    \subcaption{Video 2}
    \label{movic_2}
\end{minipage}
\vspace{1em}
\begin{minipage}{1.0\textwidth}
    \centering
    \includegraphics[width=1.0\linewidth]{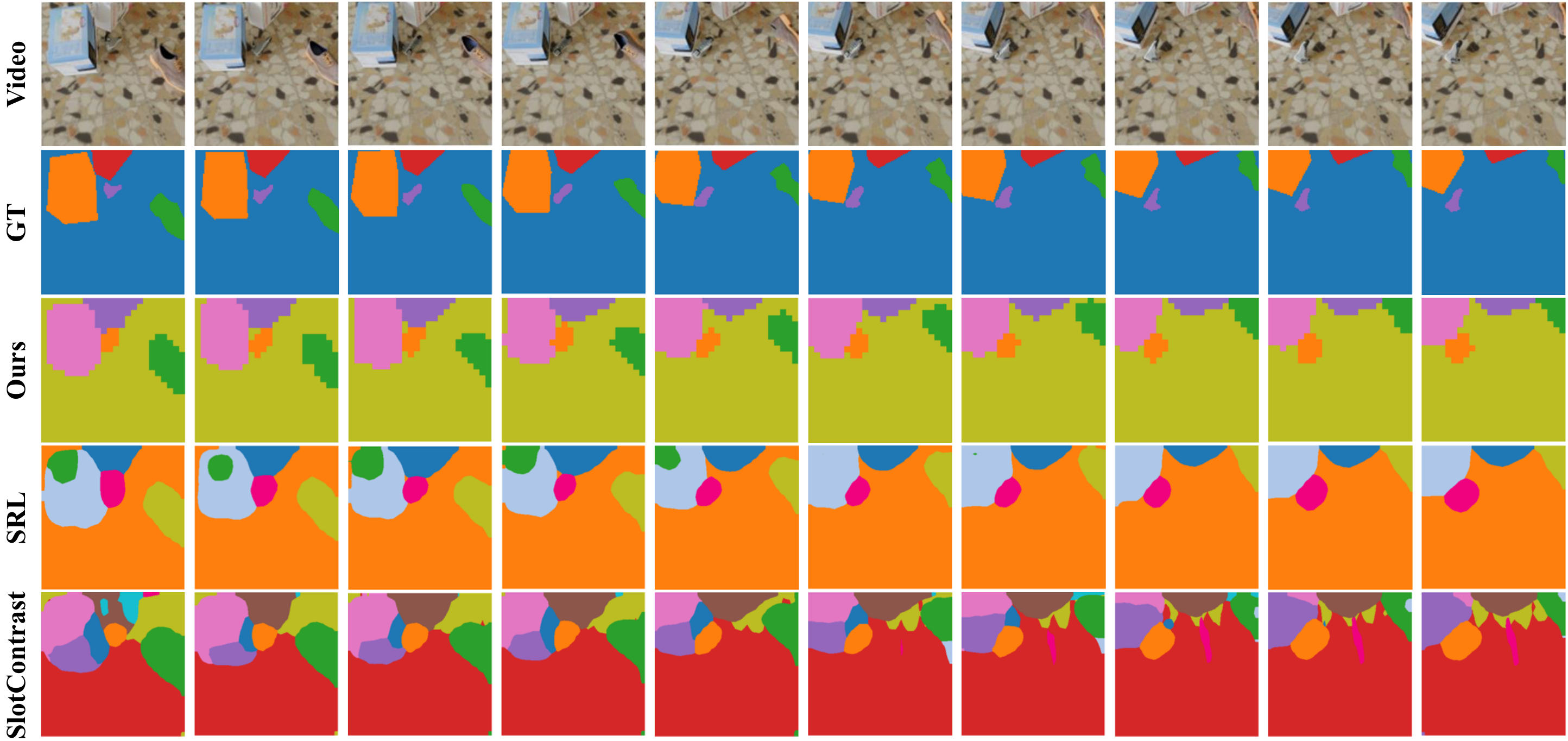}
    \subcaption{Video 3}
    \label{movic_3}
\end{minipage}
\caption{\textbf{Qualitative results on the MOVi-C dataset.}}
\label{Fig.movic_qual}
\end{figure*}
\begin{figure*}[t]
\centering
\begin{minipage}{1.0\textwidth}
    \centering
    \includegraphics[width=0.98\linewidth]{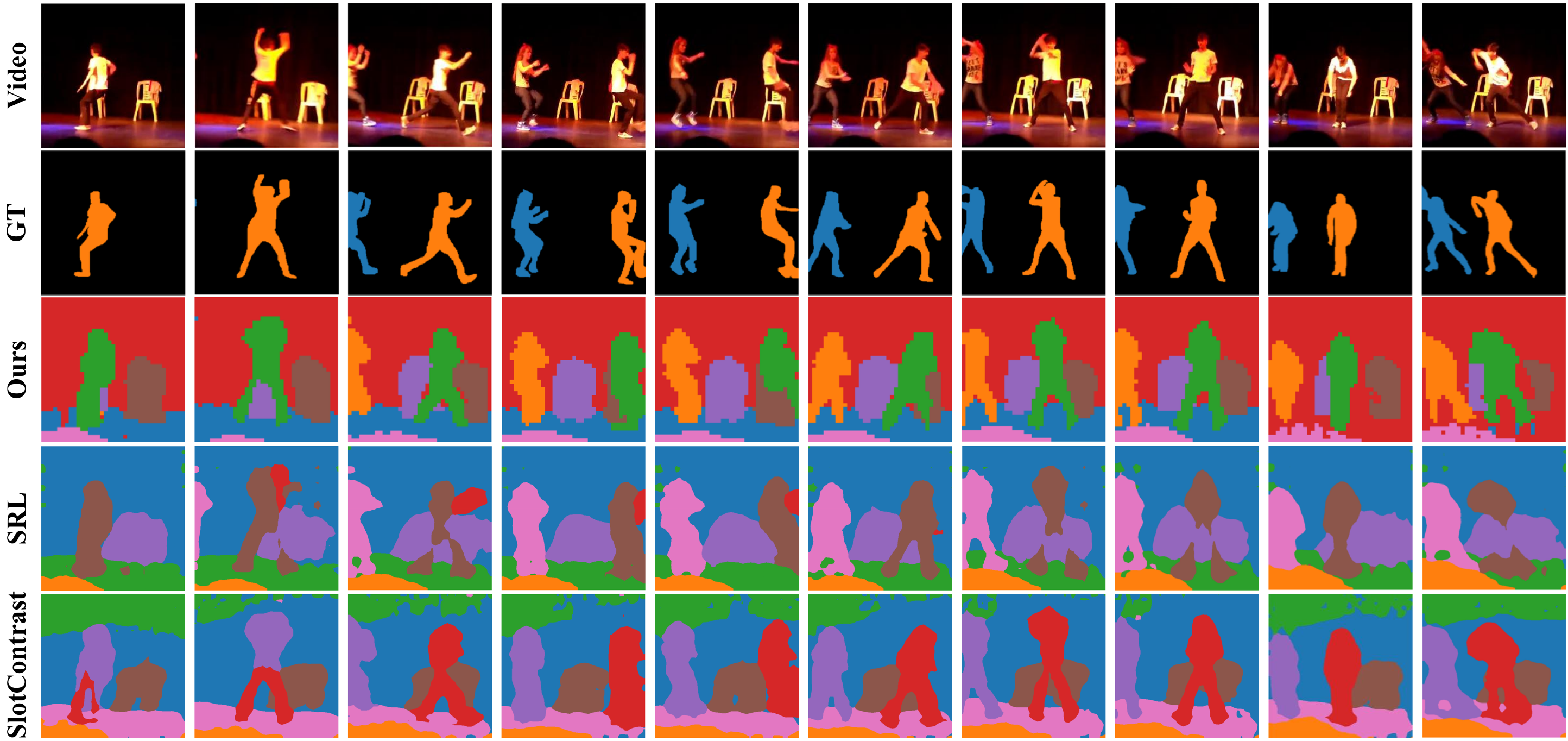}
    \subcaption{Video 1}
    \label{ytvis_1}
\end{minipage}
\vspace{1em} 
\begin{minipage}{1.0\textwidth}
    \centering
    \includegraphics[width=0.98\linewidth]{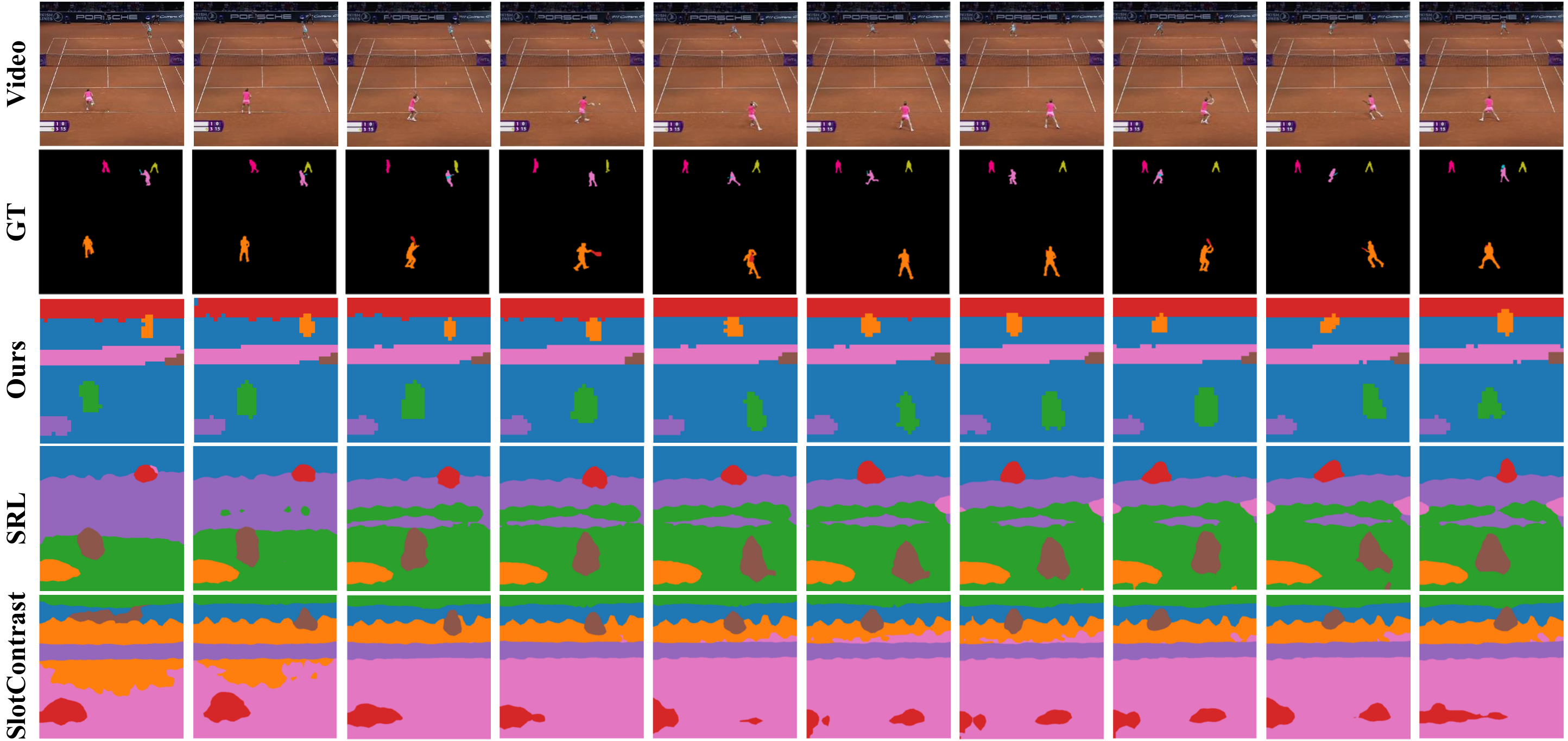}
    \subcaption{Video 2}
    \label{ytvis_2}
\end{minipage}
\vspace{1em}
\begin{minipage}{1.0\textwidth}
    \centering
    \includegraphics[width=0.98\linewidth]{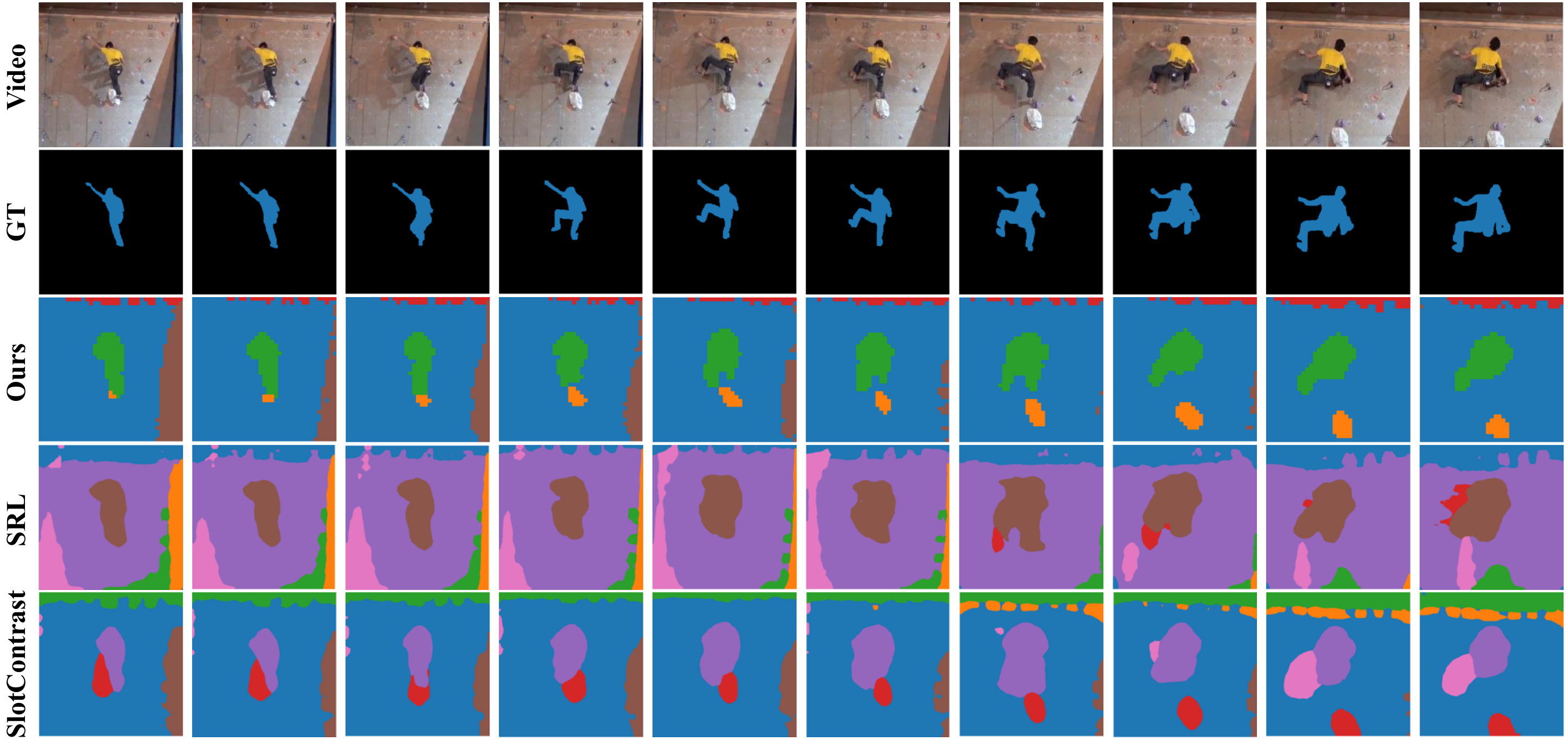}
    \subcaption{Video 3}
    \label{ytvis_3}
\end{minipage}
\vspace{-0.5cm}
\caption{\textbf{Qualitative results on the YouTube-VIS dataset.}}
\label{Fig.ytvis_qual}
\end{figure*}
We provide further qualitative comparisons on MOVi-C and the real-world YouTube-VIS 2021 benchmark, illustrating three representative sequences per dataset~(Fig.~\ref{Fig.movic_qual} and Fig.~\ref{Fig.ytvis_qual}).
On MOVi-C~(Fig.~\ref{Fig.movic_qual}), SSync produces more temporally coherent decompositions with tighter object extents than both SlotContrast and SRL.
In the first sequence~(Fig.~\ref{movic_1}), SSync assigns a semantically meaningful and stable slot to the background structure while preserving compact foreground masks, whereas SRL often misses background details and SlotContrast fragments both foreground and background regions.
In the second sequence~(Fig.~\ref{movic_2}), SSync remains robust to repetitive background textures that frequently induce spurious slot activations in prior methods, avoiding the failure mode where textured patterns are mistakenly segmented as additional objects.
In the third sequence~(Fig.~\ref{movic_3}), SSync yields object-level masks that are both less over-fragmented and sharper at boundaries, consistent with our design that leverages encoder cues for boundary refinement and decoder coherence for interior denoising.

We observe the same trends on YouTube-VIS 2021~(Fig.~\ref{Fig.ytvis_qual}), where real-world videos exhibit stronger appearance variation, non-rigid motion, and occlusions.
In the first sequence~(Fig.~\ref{ytvis_1}), SlotContrast frequently splits a single person across multiple slots in early frames, and both SlotContrast and SRL struggle to separate two visually similar chairs.
In contrast, SSync maintains stable person instances and consistently distinguishes the two separate chairs throughout the clip.
In the second sequence~(Fig.~\ref{ytvis_2}), SlotContrast fails to recover the person in the lower part of the frame, while SRL detects the person but produces a less semantically consistent grouping of the tennis-court surface; SSync preserves the person and partitions the scene with clearer, more semantically aligned boundaries.
In the final sequence~(Fig.~\ref{ytvis_3}), both SRL and SlotContrast struggle to segment the stepping stones. 
Furthermore, SRL tends to split a largely uniform background into multiple slots, and SlotContrast continues to over-fragment moving entities, whereas SSync keeps the background compact and foreground instances stable over time.
Overall, SSync more effectively distinguishes unannotated secondary objects and background elements in the visualized scenes. 
These qualitative improvements align with our strong FG-ARI and mBO performances, while large-scale quantitative validation of these specific effects in real-world videos is left for future work.

\section{Failure Case Analysis~(Limitation).}
\begin{figure*}[t]
\centering
\begin{minipage}{1.0\textwidth}
    \centering
    \includegraphics[width=1.0\linewidth]{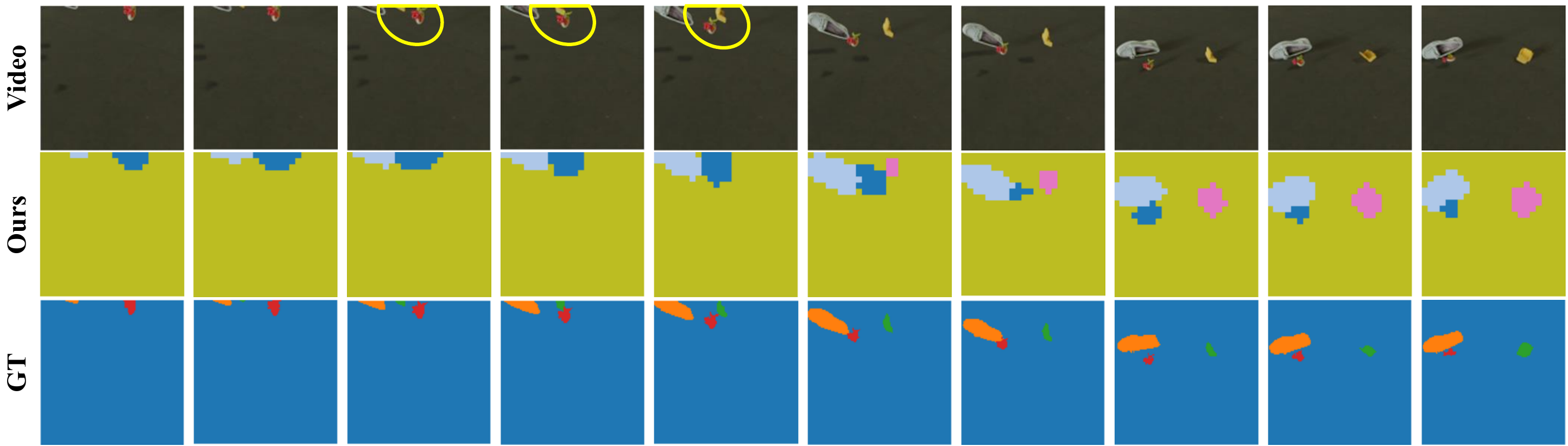}
    \subcaption{Failure modes on MOVi-C.}
    \label{Fig.failure_movic}
\end{minipage}
\vspace{1em} 
\begin{minipage}{1.0\textwidth}
    \centering
    \includegraphics[width=1.0\linewidth]{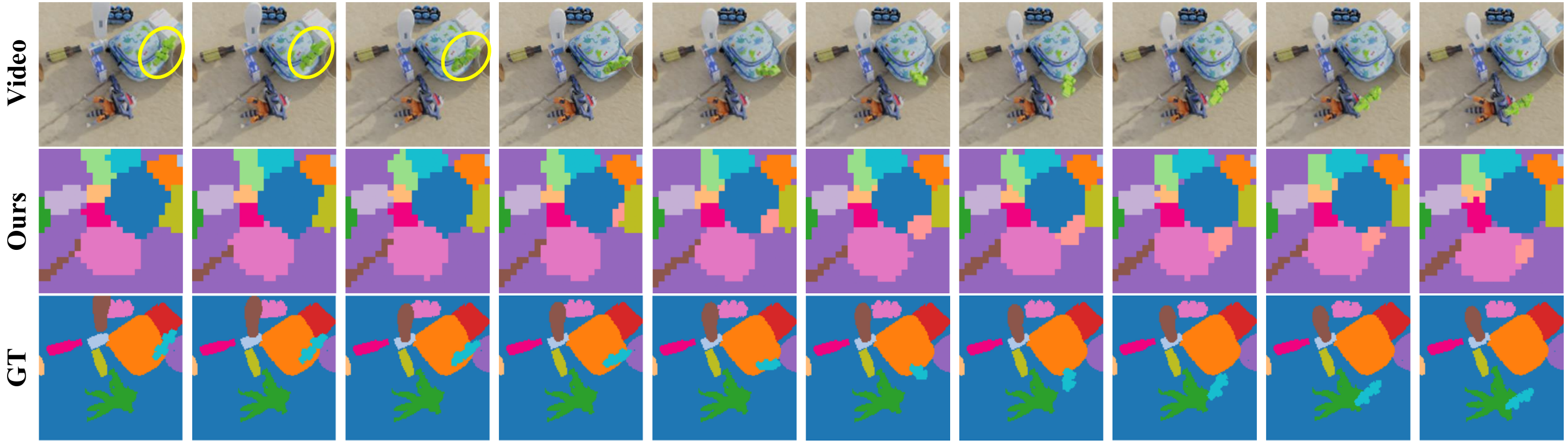} 
    \subcaption{Failure modes on MOVi-E.}
    \label{Fig.failure_movie}
\end{minipage}
\vspace{1em}
\begin{minipage}{1.0\textwidth}
    \centering
    \includegraphics[width=1.0\linewidth]{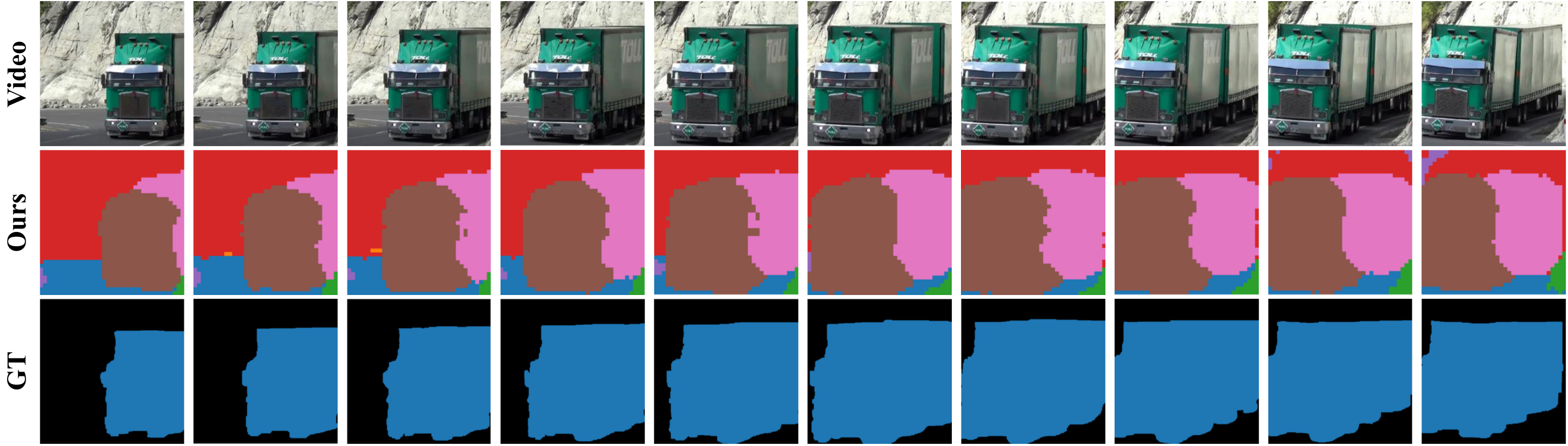} 
    \subcaption{Failure modes on YouTube-VIS 2021}
    \label{Fig.failure_ytvis}
\end{minipage}
\caption{\textbf{Failure Mode Analysis.} 
}
\label{Fig.failure}
\end{figure*}

In Fig.~\ref{Fig.failure}, we analyze representative failure cases on MOVi-C, MOVi-E, and YouTube-VIS to clarify the current limitations of SSync and to motivate future directions.
Overall, we observe two failure modes: (i) early-frame identity under-fragmentation and (ii) part-level over-fragmentation for large objects with strong intra-object variation.

First, on MOVi-C and MOVi-E datasets, we observe that under-fragmentation may occur in early frame predictions since motion cues in earlier frames may be insufficient to distinguish objects entering from similar spatial locations or overlapped, semantically similar objects.
For instance, in Fig.~\ref{Fig.failure_movic}, two small objects~(e.g., red and yellow) emerge from a similar direction and remain covered by one slot until their motion trajectories diverge; only after they separate spatially does the model consistently allocate distinct identities.
Similarly, in Fig.~\ref{Fig.failure_movie}, a green object placed in front of a bag with a similar green pattern is not immediately recognized as a separate entity, but becomes distinguishable in later frames once relative motion and spatial separation increase.
These cases suggest that incorporating bidirectional or offline temporal modeling~(e.g., using future frames during refinement) could further mitigate early-frame errors.

On YouTube-VIS, a common failure mode arises for large-scale objects whose different parts exhibit markedly different visual characteristics.
A representative example is a cargo truck~(Fig.~\ref{Fig.failure_ytvis}), where the driver's cabin and the container have substantially different textures.
Despite transitive pseudo-label merging, SSync can still assign these semantically related parts to different slots, indicating that spatio-temporal overlap alone may be insufficient when intra-object appearance variance is high.
We anticipate that addressing this limitation likely requires stronger part-to-whole grouping priors.

%% file: tab/tab_7_selection.tex
\begin{wraptable}{r}{4.6cm}
    \centering
    \vspace{-25pt}
    \scriptsize
    \caption{\textbf{Variants of boundary and interior selection mechanisms.}}
    \label{tab:selection_alternative}
    \renewcommand{\arraystretch}{1.}  
    \setlength{\tabcolsep}{2pt} 
        \begin{tabular}{ccc}
            \toprule
            Selection & FG‑ARI\,$\uparrow$ & mBO\,$\uparrow$ \\ 
            \midrule
            \rowcolor{gray!20}
            Ours & 77.1 & 38.0 \\ \hline
            \xmark & 69.0 & 30.6 \\ \hline
            Erosion & 76.4 & 36.9\\ \hline
            Entropy~(Avg) & 72.1 & 37.6 \\
            Entropy~(low 5\%) & 69.5 & 35.6 \\
            Entropy~(low 10\%) & 71.9 & 34.0 \\
            Entropy~(low 20\%) & 71.0 & 35.9 \\
            Entropy~(low 30\%) & 69.4 & 37.8 \\
            Entropy~(low 50\%) & 69.5 & 37.6 \\ 
            \bottomrule
        \end{tabular}
        \vspace{-20pt}
\end{wraptable}

%% file: tab/tab_8_additional.tex
\begin{wraptable}{r}{4.6cm}
    \centering
    \vspace{-25pt}
    \scriptsize
    \caption{\textbf{Ablation on SSync design choices.} 
    We evaluate the impact of replacing hard pseudo-labels with soft supervision~(Soft Targets/Selection) and reversing the supervision targets between the encoder and decoder~(Reversed SSync).}
    \label{tab:other_alternative}
    \renewcommand{\arraystretch}{1.}  
    \setlength{\tabcolsep}{2pt} 
        \begin{tabular}{ccc}
            \toprule
            Selection & FG‑ARI\,$\uparrow$ & mBO\,$\uparrow$ \\ 
            \midrule
            \rowcolor{gray!20}
            Ours & 77.1 & 38.0 \\ \hline
            Baseline & 69.0 & 30.6 \\ \hline
            Soft Targets & 72.9 & 34.0 \\
            + Soft Selection & 72.6 & 35.9 \\ \hline
            Reversed SSync & 67.6& 37.6 \\
            \bottomrule
        \end{tabular}
        \vspace{-20pt}
\end{wraptable}